\documentclass{article}

    \PassOptionsToPackage{numbers, compress}{natbib}

    \usepackage[preprint]{neurips_2024}

\usepackage[utf8]{inputenc} 
\usepackage[T1]{fontenc}    
\usepackage{newtxtext,newtxmath} 
\usepackage{hyperref}       
\usepackage{url}            
\usepackage{booktabs}       
\usepackage{amsfonts}       
\usepackage{nicefrac}       
\usepackage{microtype}      
\usepackage{multirow}  

\hypersetup{
    colorlinks=true,
    linkcolor=magenta,
    citecolor=blue,
    filecolor=cyan,
    urlcolor=purple
}

\usepackage[dvipsnames,svgnames,table]{xcolor}
\usepackage{graphicx}
\usepackage{amsmath}
\usepackage{booktabs}
\usepackage{multirow}
\usepackage{makecell}
\usepackage{caption}
\usepackage{subcaption}
\makeatletter
\@namedef{ver@everyshi.sty}{}
\makeatother
\usepackage{bbm}
\usepackage{wrapfig}
\usepackage{mathtools,xparse}
\usepackage{pifont}
\usepackage{algorithmic}
\usepackage[ruled,norelsize,linesnumbered]{algorithm2e}
\usepackage{enumitem}
\usepackage{pifont}
\usepackage{array}
\usepackage{scalerel,xparse}

\newcommand{\Autoref}[1]{%
  \begingroup%
  \def\algorithmautorefname{Algorithm}%
  \def\chapterautorefname{Chapter}%
  \def\subsectionautorefname{Subsection}%
  \autoref{#1}%
  \endgroup%
}

\newcommand{\SectionAutoref}[1]{%
  \begingroup%
  \def\sectionautorefname{\textcolor{blue}{\textbf{Section}}}%
  \hypersetup{linkcolor=blue}%
  \autoref{#1}%
  \endgroup%
}

\newcommand{\Appref}[1]{%
  \begingroup%
  \def\sectionrefname{\textcolor{blue}{\textbf{Appendix}}}%
  \hypersetup{linkcolor=blue}%
  \ref{#1}%
  \endgroup%
}
\usepackage{colortbl}
\usepackage{tablefootnote}
\usepackage{upgreek}
\usepackage{listings}
\usepackage{threeparttable} 

\title{Hypernetwork-Driven Model Fusion \\ for Federated Domain Generalization}

\author{%
    Marc Bartholet$^{1,}$\thanks{Equal contribution.}\,\,\enspace Taehyeon Kim$^{2, *}$\enspace Ami Beuret$^1$\enspace \textbf{Se-Young Yun$^2$\enspace Joachim Buhmann$^1$} \\
    $^1$ETH \quad $^2$KAIST AI \\
}

\begin{document}

\maketitle

\begin{abstract}

Federated Learning (FL) faces significant challenges with domain shifts in heterogeneous data, degrading performance. Traditional domain generalization aims to learn domain-invariant features, but the federated nature of model averaging often limits this due to its linear aggregation of local learning. To address this, we propose a robust framework, coined as \textbf{h}ypernetwork-based \textbf{Fed}erated \textbf{F}usion (\textit{{hFedF}}), using hypernetworks for non-linear aggregation, facilitating generalization to unseen domains. Our method employs client-specific embeddings and gradient alignment techniques to manage domain generalization effectively. Evaluated in both zero-shot and few-shot settings, {hFedF} demonstrates superior performance in handling domain shifts. Comprehensive comparisons on PACS, Office-Home, and VLCS datasets show that {hFedF} consistently achieves the highest in-domain and out-of-domain accuracy with reliable predictions. Our study contributes significantly to the under-explored field of Federated Domain Generalization (FDG), setting a new benchmark for performance in this area.

\end{abstract}


\section{Introduction}\label{sec:intro}

Federated Learning (FL) has emerged as a transformative approach, leveraging decentralized data while maintaining strict privacy standards \cite{fedavg}. FL effectively balances data protection with performance, making it particularly valuable in sensitive domains such as healthcare. In healthcare, institutions use FL for medical imaging and genetic analysis while keeping patient data secure \cite{fl_health1,fl_health2}. 
Despite these advantages, FL faces significant challenges when data is not independently and identically distributed (non-iid) across clients, leading to degraded performance \cite{feddgsurvey}.

\begin{figure}[!t]
    \centering
    \begin{subfigure}{0.6\textwidth}
        \includegraphics[width=\linewidth]{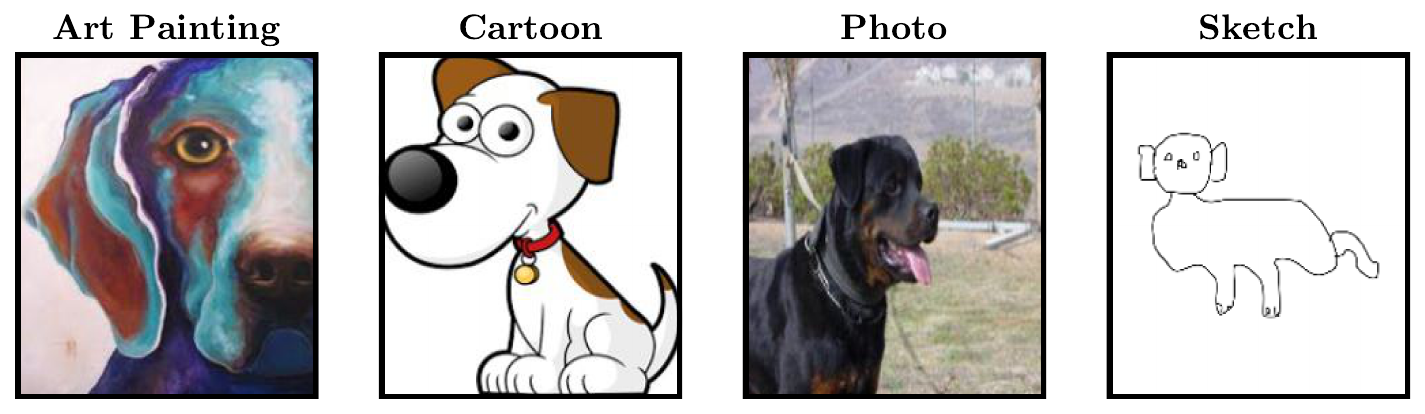}
        \caption{Example dog images}     \label{fig:data}
    \end{subfigure}
    \begin{subfigure}{0.38\textwidth}
        \includegraphics[width=\linewidth]{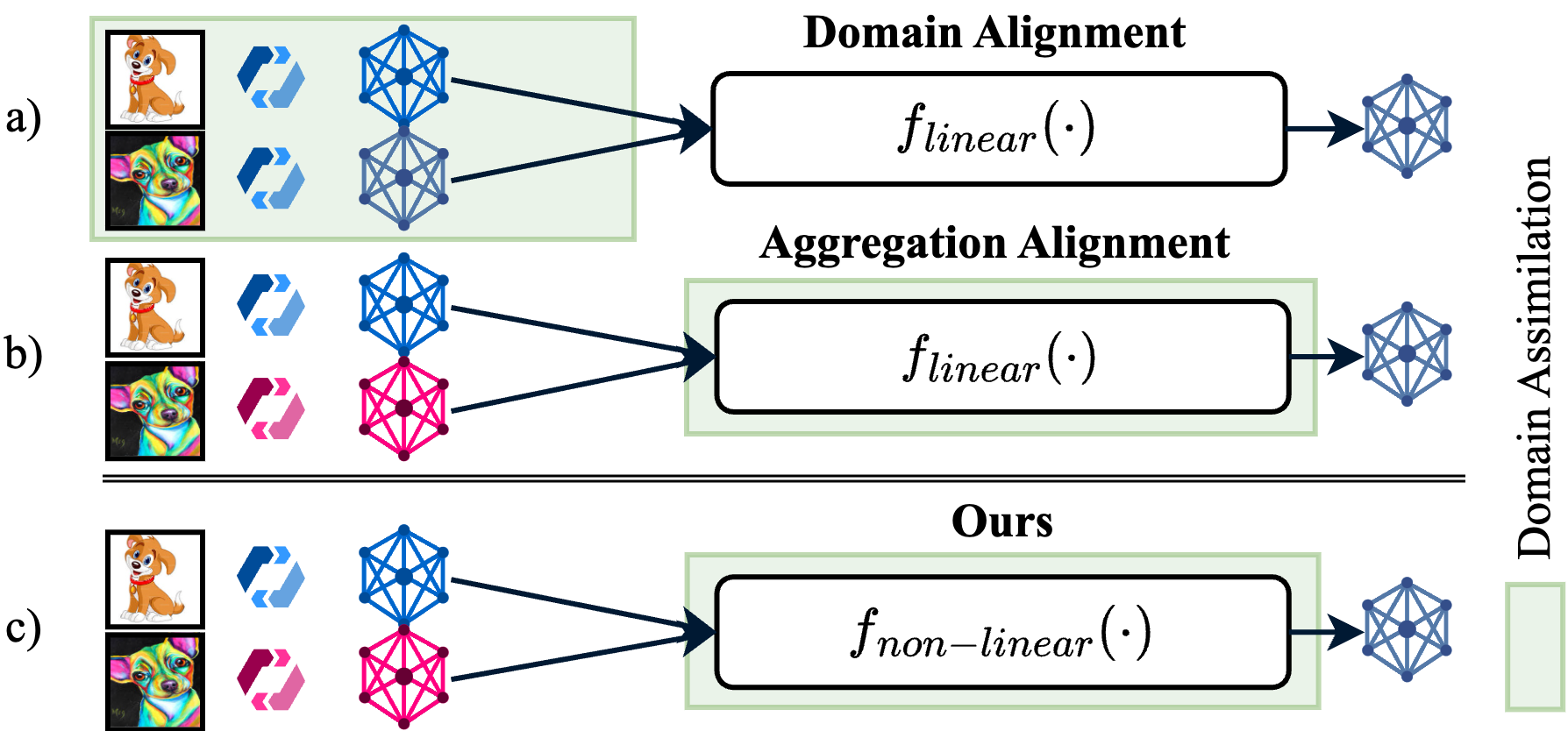}
        \caption{Overview of FDG} \label{fig:intro}
    \end{subfigure}
    \vspace{-5pt}
    \caption{(a) Examples of dog images from different domains in PACS dataset \cite{pacs}. (b) An overview of FDG approaches: \textbf{a)} Domain-invariant feature extraction (domain alignment). \textbf{b)} Linear aggregation of models (aggregation alignment). \textbf{c)} Our method: non-linear hypernetwork fusion.}
    \vspace{-10pt}
\end{figure}

A major challenge in real-world FL applications is the divergence of domain-specific data, known as \textit{domain shift} (\Autoref{fig:data}). For example, autonomous driving systems, which depend on robust object detection, must handle domain variations caused by different weather conditions \cite{fl_driving5, fl_driving6}. Federated Domain Generalization (FDG) aims to enable FL systems to generalize to unseen domains. Traditional Domain Generalization (DG) strategies focus on aligning feature distributions across multiple domains to learn domain-invariant features, which typically operate at the latent level during batch-level training that includes data from multiple domains. However, FDG cannot access domain data from different clients due to privacy constraints, making it challenging to apply these traditional DG techniques directly in a federated setup \cite{zhang2023}.

Current FDG efforts typically focus on \textit{federated domain alignment}, aiming to find domain-invariant representations during client-side training through adversarial training \cite{wang2022, peng2019, fedadg} and regularization over feature representations and a conditional mutual information \cite{fedsr} (\autoref{fig:intro}\textcolor{magenta}{-a}). Additionally, some approaches address domain shift at the aggregation stage by aligning model weights to reflect different domains \cite{yuan2023, chen2022, yang2023, zhang2023} (\autoref{fig:intro}\textcolor{magenta}{-b}). While these methods train client models on their local domain data and then aggregate them, the aggregation process itself remains largely linear, relying on the FedAvg-based weight averaging strategy. This linearity can oversimplify the complex nature of domain shifts \cite{feddg_benchmarks, feddgsurvey}, potentially diminishing the richness of local representations and compromising in-distribution (\textit{id}) performance for the sake of out-of-distribution (\textit{ood}) generalization. Therefore, there is a pressing need for more sophisticated approaches in FDG to better capture and address these complexities.


This paper presents a novel framework for FDG, termed as \textit{hFedF} (\textit{\textbf{h}ypernetwork-based \textbf{Fed}erated \textbf{F}usion}), utilizing a hypernetwork as a server-side aggregator to leverage collective knowledge from client models through adaptive parameter-sharing (\autoref{fig:intro}\textcolor{magenta}{-c}). A hypernetwork \cite{hnet} is a neural network generating parameters for client models based on unique client embeddings. Our framework employs a hypernetwork to implicitly share client weight parameters, creating a quasi-global model for all clients. While federated hypernetworks have shown promise under specific conditions (e.g., label heterogeneity, personalization) \cite{pfedhn, pefll, pfedla, hyperfl_case}, to the best of our knowledge, our work is the first to explore the use of hypernetworks for FDG, marking a significant step forward in this area of research. Our key contributions are as follows:

\begin{itemize}
    \item We introduce \textit{\textbf{h}ypernetwork-based \textbf{Fed}erated \textbf{F}usion} (\textit{hFedF}), a novel hypernetwork framework as a non-linear model aggregator. hFedF maintains the privacy through client-specific embeddings as a client identifier within a hypernetwork (\SectionAutoref{sec:method}).

    \item We show that applying hypernetworks in FDG faces challenges with stability, convergence, and synchronous updates due to differing domain characteristics. To address these, we introduce \textsc{GradAlign} and Exponential Moving Average \textsc{(EMA)} techniques for stable and effective hypernetwork parameter training in FDG (\SectionAutoref{sec:method}).

    \item We provide a comprehensive analysis of how hypernetworks differ from linear aggregation methods in training dynamics. Our extensive empirical evaluation demonstrates that our approach consistently outperforms existing benchmarks on challenging multi-domain datasets (PACS \cite{pacs}, Office-Home \cite{officehome}, and VLCS \cite{vlcs}) in federated scenarios (\SectionAutoref{sec:exp}).

\end{itemize}

\vspace{-5pt}
\section{Related Work}
\vspace{-5pt}
\subsection{Federated Learning and Domain Generalization} \label{subsec:relwork:fl_dg}

\textbf{Federated Learning (FL)} has predominantly focused on addressing challenges arising from skewed label heterogeneity among clients. The \textit{FedAvg} algorithm, which trains a global model by averaging clients' weight parameters, remains a benchmark for state-of-the-art FL algorithms due to its simplicity and smooth loss landscape \cite{fedavg, feddg_benchmarks, hmoe, feddf}. However, with \textit{non-iid} data, the average of local optima may drift from the global optimum. To mitigate this, approaches like {FedProx} \cite{fedprox} introduce an $L_2$ regularization term to the local loss, while \citet{scaffold} propose global and local control variates to better estimate update directions for both server and clients \cite{feddyn, fednova}.

Despite advancements in handling label heterogeneity, domain generalization remains underexplored in the FL community. \textbf{Domain Generalization (DG)} aims to enhance model generalization to unseen domains, which is crucial for FL systems. Traditional DG techniques, such as domain alignment \cite{li2018}, meta-learning \cite{li2017}, and regularization strategies \cite{wang2019}, have shown significant success in centralized settings but are challenging to adapt to FL. Recent efforts in Federated Domain Generalization (FDG) have introduced promising approaches. For instance, {FedDG} \cite{liu2021feddg} exchanges image distributions in the frequency space across clients and uses explicit regularization to promote domain-independent features. Other methods leverage adversarial training techniques \cite{wang2022, peng2019, fedadg} or enforce probabilistic feature representations with regularizers \cite{fedsr}. Furthermore, optimizing aggregation weights to assimilate contributions from diverse domains has been explored by \citet{yuan2023, chen2022, yang2023, zhang2023}. Recent advanced methods like FedGMA \cite{fedgma} implicitly adjust aggregation weights, and \citet{tian2023} align local gradients to uncover consistent patterns shared among clients.

\subsection{Hypernetwork}
Hypernetworks (a.k.a. meta-networks) \cite{hnet} have garnered significant attention in multi-domain learning \cite{da_hnet, hmoe, hyperformer} for their ability to facilitate joint learning by sharing knowledge across domains. For instance, {HMOE} \cite{hmoe} proposes a hypernetwork-based Mixture of Experts for domain generalization that learns the embedding space of the input and minimizes the divergence between predicted and established embeddings. The application of hypernetworks within a federated framework is relatively new, with only a limited number of studies addressing this topic. Building on the success of hypernetworks in centralized setups, \citet{pfedhn} introduced {pFedHN}, the first hypernetwork-based alternative to established FL techniques. This approach generates client-specific parameters asynchronously based on client embeddings. Similarly, \citet{pfedla} employed a hypernetwork to output a weight matrix for each client, identifying mutual contributions at the layer granularity.

Despite these advancements, the potential of hypernetworks in Federated Domain Generalization (FDG) remains underexplored. Most existing works \cite{hyperfl_case, pefll} have focused on using hypernetworks for generating highly personalized models in scenarios with label heterogeneity, without addressing the challenges posed by domain shifts.
\vspace{-5pt}
\section{Problem statement}
\vspace{-5pt}
\paragraph{General FL framework} In a typical FL setup, $N$ clients each have a local dataset $\mathcal{D}_i = \{(x_i^j, y_i^j)\}$ from their respective data distributions $\mathcal{P}_i$. Each client trains a local model $\mathcal{M}_i$ with parameters $\varphi_i$. FL aims to minimize the global objective function, a weighted average of local objective functions: $ \min_{\{\varphi_i\}_{i=1}^N} \mathcal{L}(\{\varphi_i\}_{i=1}^N) := \frac{1}{\sum_i \gamma_i}\sum^N_{i=1} \gamma_i \mathcal{L}_i(\varphi_{i})$. The local objective function for each client is $\mathcal{L}_i(\varphi_i) = \mathbb{E}_{(x,y) \sim \mathcal{P}_i}[\ell_i(y; x, \varphi_i)] \approx \frac{1}{|\mathcal{D}_i|} \sum_{j=1}^{|\mathcal{D}_i|} \ell_i(y_i^j; x_i^j, \varphi_i)$ where $\ell_i: \mathcal{Y} \times \mathcal{Y} \rightarrow \mathbb{R}_+$ is the local loss function.

\paragraph{Specific FDG challenges}
FDG extends the typical FL setup by addressing domain shifts and improving generalization to unseen domains. Unlike traditional FL, which focuses on minimizing the expected loss on training data (\textit{id}-accuracy), FDG also aims to optimize performance on unseen target domains (\textit{ood}-accuracy). This involves several specific challenges:

\begin{itemize}
    \item \textbf{Heterogeneous domain shifts:} FDG must handle significant variations in data distributions across domains, which are often non-trivial and multifaceted.
    \item \textbf{Generalization under isolation:} Each client must generalize knowledge from its local domain to an unseen target domain without access to other clients' data, making it difficult to capture the diversity needed for robust models.
    \item \textbf{Effective knowledge aggregation:} Traditional linear aggregation methods struggle to combine knowledge from diverse domains effectively due to their inability to capture complex, non-linear relationships between data distributions.
\end{itemize}

\begin{wraptable}{r}{0.44\textwidth}\small
\centering
\vspace{-13pt}
\caption{Comparison of previous literature's \cite{hyperfl_case, pefll} and our hypernetwork in FDG.\label{tab:hyp_diff}}
\vspace{-3pt}
\addtolength{\tabcolsep}{-3.5pt}
\begin{tabular}{l|c|c} 
\toprule
Properties & Literature \cite{hyperfl_case, pefll} & Ours \\ 
\midrule
Stability & $\times$ & $\surd$ \\
Convergence & $\times$ & $\surd$ \\
Synchronous Update & $\times$ & $\surd$ \\
\bottomrule
\end{tabular}
\vspace{-10pt}
\end{wraptable}

\paragraph{Hypernetwork-driven aggregation} 
To address the challenges of FDG, we propose using a hypernetwork as a non-linear aggregator in the federated architecture. Hypernetworks, also known as meta-networks, can effectively capture complex relationships between domains, enhancing the model's ability to generalize to unseen target domains. Prior research has demonstrated that meta-networks are powerful in multi-domain learning as they facilitate knowledge sharing across domains, enabling better generalization \cite{hnet} and personalization \cite{ruiz2023hyperdreambooth}. However, integrating hypernetworks into FDG introduces several technical considerations (\autoref{tab:hyp_diff}):

\begin{itemize}
    \item \textbf{Stability:} Ensuring stable training of hypernetworks is challenging due to the diverse and heterogeneous data distributions across clients. Stability issues can lead to significant performance degradation in FDG scenarios.
    
    \item \textbf{Convergence:} Achieving convergence in a global model is complex as it requires effective aggregation of knowledge from all clients, each with different domain-specific distributions.
    
    \item \textbf{Synchronous updates:} Implementing synchronous updates is essential to maintain consistency and performance across clients. However, this is particularly challenging in a federated hypernetwork scenario where data distributions and training dynamics vary significantly.
\end{itemize}

\begin{figure}[t]
\begin{center}
\centerline{\includegraphics[width=\textwidth]{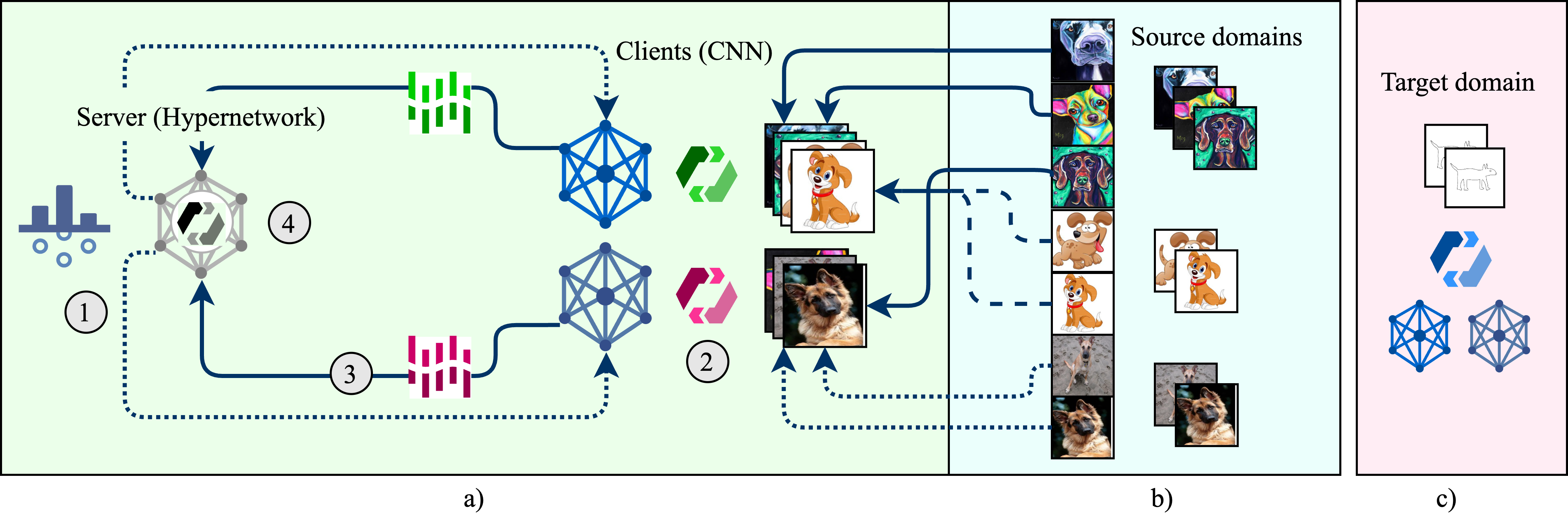}}
\caption{\textbf{a)} Our hFedF framework with source domains (\protect\SectionAutoref{sec:method}). \textbf{b)} Domain splitting strategy: Visualization of splitting 3 domains across 2 clients, each holding samples from 2 domains ($d=2$). The largest domain is split into more parts to meet constraints. \textbf{c)} At inference, generalization is evaluated on an unseen target domain, and personalization is assessed on a held-out subset of the local source domain data.}
\vspace{-20pt}
\label{fig:algo}
\end{center}
\end{figure}

\vspace{-5pt}
\section{Method}\label{sec:method}
\subsection{Hypernetwork-driven federated optimization}

Our methodology, \textit{hFedF} (Hypernetwork-based Federated Fusion), integrates a hypernetwork $\boldsymbol{h}$ to dynamically aggregate client-specific models in a non-linear fashion (\Autoref{fig:algo}). At the onset of each communication round, the server utilizes $\boldsymbol{h}$ to generate and distribute tailored model parameters to each client (\textbf{1} in \Autoref{fig:algo}). Clients then leverage their local datasets to update these parameters over a series of iterations. Post-update, the clients compute the differential gradients by comparing the updated parameters with the initial ones received from the server (\textbf{2} in \Autoref{fig:algo}). These gradients are synchronized and aggregated on the server based on their alignment with the primary update direction (\textbf{3} in \Autoref{fig:algo}). The culmination of this process sees the server refining $\boldsymbol{h}$ by integrating a moving average of these updates to enhance the model’s stability and convergence (\textbf{4} in \Autoref{fig:algo}).

\begin{figure}[t]
\begin{algorithm}[H]
\DontPrintSemicolon
\caption{\textit{hFedF}} \label{algo:fedtrain}
\begin{algorithmic}[1]
\INPUT: $T$ (communication rounds), $E$ (local epochs), $\alpha$ (server learning rate), $\mu$ (client learning rate), $\mathcal{D}_i$ (local data of client $i$), $\mathcal{S}$ (set of clients)
\STATE {\bfseries Initialize:} $\theta^0, \nu^0$ \tcp*{Initialize hypernetwork \& client embeddings}
\FOR{$t \leftarrow 0, \dots, T-1$}
    \STATE \textcolor{darkgray}{/* Client-side updates to train local models */}
    \FOR{each client $i \in \mathcal{S}$ {\bfseries in parallel}}
        \STATE $\varphi_i^{t,0} \gets h(\theta^t, \nu^t[i])$ \tcp*{Initialize local model parameters}
        \FOR{$e \leftarrow 0, \dots, E-1$}
            \STATE $\tilde{\varphi}_i^{t,e+1} \gets \textsc{ClientUpdate}(\tilde{\varphi}_i^{t,e}; \mathcal{D}_i, \mu)$
        \ENDFOR
        \STATE $\Updelta\varphi_i^t \gets \tilde{\varphi}_i^{t,E} - \varphi_i^{t,0}$ \tcp*{Update local model}
    \ENDFOR
    \STATE \textcolor{darkgray}{/* Server-side updates to aggregate knowledge */}
    \FOR{each client $i \in \mathcal{S}$}
        \STATE $g_{\theta,i}^t, g_{\nu,i}^t \gets \nabla_{\theta} h(\theta^t, \nu^t[i])^\mathsf{T} \cdot \Updelta\varphi_i^t, \nabla_{\nu} h(\theta^t, \nu^t[i])^\mathsf{T} \cdot \Updelta\varphi_i^t$ \tcp*{for hypernetwork}
        \STATE $\tilde{\gamma}_{\theta,i}^t, \tilde{\gamma}_{\nu,i}^t \gets \textsc{GradAlign}(g_{\theta,i}^t, \mathcal{S}), \textsc{GradAlign}(g_{\nu,i}^t, \mathcal{S})$ \tcp*{for client embedding}
    \ENDFOR
    \STATE $g_\theta^t, g_\nu^t \gets \sum_{i \in \mathcal{S}} \tilde{\gamma}_{\theta,i}^t \cdot g_{\theta,i}^t, \sum_{i \in \mathcal{S}} \tilde{\gamma}_{\nu,i}^t \cdot g_{\nu,i}^t$\tcp*{Aggregation}
    \STATE $\theta^{t+1}, \nu^{t+1} \gets \theta^t - \alpha \cdot g_\theta^t, \nu^t - \alpha \cdot g_\nu^t$ \tcp*{Update}
    \STATE \textsc{EMA} $\{\theta^{t+1}, \nu^{t+1}\} \gets \textsc{EMA}(\{\theta^{t+1}, \nu^{t+1}\}, t)$ \tcp*{Apply EMA for stability}
\ENDFOR
\end{algorithmic}
\end{algorithm}
\end{figure}

The hypernetwork, parameterized by $\theta$, processes embeddings $\nu \in \mathbb{R}^{L \times N}$ to generate client-specific parameters $\varphi_{i} = \boldsymbol{h}(\theta, \nu[i]),\,\, \forall i \in \{1, \ldots, N\}$. The global optimization problem is expressed as:
\[
\min_{\theta, \nu} \left\{\mathcal{L}(\theta, \nu) := \frac{1}{\sum_i \gamma_i}\sum^N_{i=1} \gamma_i \mathcal{L}_i(h(\theta, \nu[i]))\right\}.
\]
with gradients calculated via: $\nabla_{\theta, \nu}\mathcal{L}_i(\varphi_i) = \nabla_{\theta, \nu} \varphi_i^T \cdot \nabla_{\varphi_i} \mathcal{L}_i(\varphi_i)$, facilitating the propagation of gradients back to the hypernetwork's parameters. Utilizing a hypernetwork architecture offers several key benefits in FL:
\begin{itemize}
    \item \textbf{Communication efficiency}: Communication is limited to local model parameters, not the larger hypernetwork itself, allowing for complex server-side models without added overhead.
    \item \textbf{Privacy protection}: Compromising a client model does not expose information about other clients, as individual embeddings remain on the server and are not interpretable.
\end{itemize}

This hypernetwork-driven framework not only reduces communication demands typical in federated environments but also protects privacy. Despite these advantages, applying hypernetworks in FDG poses challenges including stability and synchronization across diverse client domains, which are addressed in following sections.



\subsection{Addressing challenges of hypernetworks in FDG}

\textbf{Synchronous updates via \textsc{GradAlign}}\, Effective synchronous updates is essential to maintain consistency and performance across clients. This is particularly challenging in a federated multi-domain setting where data distributions and training dynamics vary significantly. To mitigate \textit{client drift}, we introduce a non-parametric gradient alignment technique. The server applies the chain rule to each approximate gradient $\Updelta\varphi_i$ to obtain the proposed update direction $g_i$. The average update direction $g_{\text{avg}}$ is then calculated, and the alignment of each gradient $g_i$ with $g_{\text{avg}}$ is assessed using cosine similarity, yielding preliminary aggregation weights $\gamma_i$:
\[
\gamma_i = \frac{g_{\text{avg}} \cdot g_i}{||g_{\text{avg}}|| \ ||g_i||}, \text{ where } g_{\text{avg}} = \frac{1}{N} \sum_{i=1}^N g_i.
\]
These weights are transformed into a probability distribution using the softmax function: $\tilde\gamma_i = \frac{e^{-\gamma_i}}{\sum_{j=1}^N e^{-\gamma_j}}$.
The final hypernetwork gradient is determined through a linear combination of local gradients, considering these weights. 

\textbf{Stability via \textsc{EMA} for hypernetwork weight parameters}\, Ensuring stable training of hypernetworks is challenging due to the diverse and heterogeneous data distributions across clients. Stability issues can lead to significant performance degradation in FDG scenarios. To address this, we employ Exponential Moving Average (EMA) regularization. The smoothed model $\theta_{\textit{EMA}}^t$ is computed as a weighted average of the current model $\theta^t$ and the previous smoothed model $\theta_{\textit{EMA}}^{t-1}$: $\theta_{\textit{EMA}}^t = \alpha \theta^t + (1-\alpha) \theta_{\textit{EMA}}^{t-1}$. The value of $\alpha$ is mainly set between 0.75 and 0.95. More descriptions are in \textcolor{blue}{Appendix}~\Appref{appendix:hyperparams}

\textbf{Convergence}\, Achieving convergence in FDG is challenging due to the non-linear and complex relationships between client models with different domain-specific distributions. The combination of \textsc{EMA} and \textsc{GradAlign} addresses these challenges by enhancing the stability and synchronization of updates. \textsc{EMA} regularizes the training process, smoothing out fluctuations and maintaining responsiveness to recent changes. \textsc{GradAlign} ensures that client updates are well-aligned and contribute effectively to the global model. Together, these techniques lead to faster and more robust convergence, ensuring the model performs well across diverse domains.

\textbf{Optimization of client embeddings}\, Effective client embeddings are crucial for personalizing the hypernetwork's outputs to individual data distributions within a federated network. We evaluated various strategies for generating these embeddings, including using randomized embeddings and deriving them from a pretrained autoencoder (\textcolor{blue}{Appendices}~\Appref{appendix:embeddings} and \Appref{appendix:embeddings2}). Our findings indicate that dynamically learning embeddings from scratch during the federated learning process is the most effective method. This approach allows the hypernetwork to continuously refine and adapt embeddings, ensuring they are optimally tailored to the unique characteristics of each client's data.
\vspace{-10pt}
\section{Experiments}\label{sec:exp}
\vspace{-5pt}
\subsection{Setup}

\textbf{Datasets}\, We evaluate our approach on three datasets: PACS \cite{pacs}, Office-Home \cite{officehome}, and VLCS \cite{vlcs}. PACS contains 9,991 images from four domains: art painting, cartoon, photo, and sketch, with a classification task involving seven classes. Office-Home consists of 15,500 images of everyday objects across four domains: art, clipart, product, and real, featuring 65 classes. VLCS includes data from four domains: Pascal VOC 2007 \cite{pascal-voc-2007}, LabelMe \cite{russell2008labelme}, Caltech101 \cite{caltech101}, and SUN09 \cite{sun09}, presenting a variety of objects and scenes. These datasets, with their increasing classification difficulty and diverse domain representations, provide a comprehensive evaluation of the generalization capabilities.

\textbf{Domain heterogeneity and evaluation}\, Following {DomainBed} \cite{domainbed} and FedSR \cite{fedsr} guidelines, we adopt a \textit{leave-one-domain-out} validation approach. All but one domain are used as source domains ($\mathcal{Z}_{\text{src}} = \{Z_k\}_{k=1}^{K-1}$), with the remaining domain serving as the target domain ($\mathcal{Z}_{\text{trg}} = \{Z_K\}$). Each client retains 10\% of its source data as a validation set ($\mathcal{D}_{i, \text{val}}$) to assess \textit{id}-accuracy, while the target domain data ($\mathcal{D}_K$) is used to evaluate \textit{ood}-accuracy. To address domain heterogeneity, we propose a novel domain allocation strategy controlled by the hyperparameter $d$, which defines the number of distinct domains included in each client's training data. Domains are partitioned to clients based on $d$.

\textbf{Client setup and model architectures}\, We set the number of clients $N$ to the number of source domains ($|\mathcal{Z}_{\text{src}}|$). Following \citet{pfedhn}, the embedding dimension $L$ is set to $\lfloor 1 + \frac{N}{4} \rfloor$, with additional results for higher dimensions and more clients provided in \textcolor{blue}{Appendix}~\Appref{appendix:embeddings}. The hypernetwork is a multi-layer perceptron with an embedding layer followed by three hidden layers of 50 neurons each, ending in a multi-head structure that maps to the dimensions of the client model layers. The client model uses a convolutional neural network .

\textbf{Training details and baseline comparisons}\, We test all combinations of source and target domains, averaging results across three different seeds. Each algorithm employs the same client model and is trained for 200 communication rounds, with 2 local epochs per round and a batch size of 64. Hyperparameters are tuned per dataset according to reference papers. Detailed results and additional setup information are provided in the \textcolor{blue}{Appendices}~\Appref{appendix:experiments}, \Appref{appendix:moreepochs} and \Appref{appendix:morerounds}. For baseline comparisons, we evaluate \textit{hFedF} against several prominent FL benchmarks, including \textit{FedAvg} \cite{fedavg}, a foundational FL algorithm that averages model updates from clients; \textit{FedProx} \cite{fedprox}, an extension of FedAvg that adds a proximal term to address data heterogeneity; and \textit{pFedHN} \cite{pfedhn}, a personalized FL method using hypernetworks for client-specific models. We also compare against FDG benchmarks such as \textit{FedSR} \cite{fedsr}, which focuses on regularizing feature representations to mitigate domain shifts, and \textit{FedGMA} \cite{fedgma}, which adjusts aggregation weights to handle domain heterogeneity. Additionally, we include two non-federated baselines: \textit{Central}, where all domains' data is aggregated and trained on a single model, and \textit{Local-Only}, where each client trains independently on its local data without any communication. 

\begingroup
\setlength{\tabcolsep}{6pt} 
\renewcommand{\arraystretch}{1.17}
\begin{table}[t]
\caption{Averaged accuracy across domains on PACS \cite{pacs}, Office-Home \cite{officehome}, and VLCS \cite{vlcs}. The best accuracy along a column is marked in \textbf{bold}, with the exception of non-federated benchmarks. The arrow (\textcolor{red}{$\downarrow$}, \textcolor{ForestGreen}{$\uparrow$}) shows the comparison to the FedAvg.}
\label{tab:main}
\vspace{-5pt}
\begin{center}
\resizebox{\textwidth}{!}{%
\begin{tabular}{lcccccccccccc} 
\toprule
 & \multicolumn{4}{c}{PACS} & \multicolumn{4}{c}{Office-Home} & \multicolumn{4}{c}{VLCS} \\
\cmidrule{2-13}
 & \multicolumn{2}{c}{$d = 1$} & \multicolumn{2}{c}{$d = 2$} & \multicolumn{2}{c}{$d = 1$} & \multicolumn{2}{c}{$d = 2$} & \multicolumn{2}{c}{$d = 1$} & \multicolumn{2}{c}{$d = 2$} \\ 
\cmidrule{2-13}
 & $\mu_\text{\textit{id}}$ & $\mu_\text{\textit{ood}}$ & $\mu_\text{\textit{id}}$ & $\mu_\text{\textit{ood}}$ & $\mu_\text{\textit{id}}$ & $\mu_\text{\textit{ood}}$ & $\mu_\text{\textit{id}}$ & $\mu_\text{\textit{ood}}$ & $\mu_\text{\textit{id}}$ & $\mu_\text{\textit{ood}}$ & $\mu_\text{\textit{id}}$ & $\mu_\text{\textit{ood}}$ \\ 
\midrule
 Local-only & 76.4 & 35.8 & 71.5 & 43.3 & 47.7 & 17.2 & 40.5 & 19.7 & 69.9 & 43.6 & 63.2 & 50.0 \\ 
\midrule
 FedAvg \cite{fedavg} & 70.1 & 51.3 & 75.5 & 53.2 & 47.8 & 29.1 & 51.3 & 29.4 & 65.6 & \underline{\textbf{58.0}} & 65.6 & 57.7 \\
 \midrule
 FedProx \cite{fedprox} & 69.9 \textcolor{red}{$\downarrow$} & 50.7 \textcolor{red}{$\downarrow$} & 75.1 \textcolor{red}{$\downarrow$} & 53.0 \textcolor{red}{$\downarrow$} & 48.6 \textcolor{ForestGreen}{$\uparrow$} & 29.1 \textcolor{ForestGreen}{$\uparrow$} & 51.5 \textcolor{ForestGreen}{$\uparrow$} & 29.7 \textcolor{ForestGreen}{$\uparrow$} & 65.4  \textcolor{red}{$\downarrow$} & 56.2 \textcolor{red}{$\downarrow$} & 64.6 \textcolor{red}{$\downarrow$} & 56.7 \textcolor{red}{$\downarrow$} \\
 pFedHN \cite{pfedhn} & 65.2 \textcolor{red}{$\downarrow$} & 48.2 \textcolor{red}{$\downarrow$} & 72.2 \textcolor{red}{$\downarrow$} & 50.6  \textcolor{red}{$\downarrow$} & \underline{\textbf{49.9}} \textcolor{ForestGreen}{$\uparrow$} & 21.1 \textcolor{red}{$\downarrow$} & 45.1 \textcolor{red}{$\downarrow$} & 23.6 \textcolor{red}{$\downarrow$} & 63.4 \textcolor{red}{$\downarrow$} & 56.2 \textcolor{red}{$\downarrow$} & 63.9 \textcolor{red}{$\downarrow$} & 57.0 \textcolor{red}{$\downarrow$} \\ 
 FedSR \cite{fedsr} & 69.2 \textcolor{red}{$\downarrow$} & 48.9 \textcolor{red}{$\downarrow$} & 73.8 \textcolor{red}{$\downarrow$} & 51.6 \textcolor{red}{$\downarrow$} & 48.1 \textcolor{red}{$\downarrow$} & 29.6 \textcolor{ForestGreen}{$\uparrow$} & 51.5 \textcolor{ForestGreen}{$\uparrow$} & 29.8 \textcolor{ForestGreen}{$\uparrow$} & 66.0 \textcolor{ForestGreen}{$\uparrow$} & 57.2 \textcolor{red}{$\downarrow$} & 65.2 \textcolor{red}{$\downarrow$} & 57.4 \textcolor{red}{$\downarrow$} \\
 FedGMA \cite{fedgma} & 68.7 \textcolor{red}{$\downarrow$} & 48.9 \textcolor{red}{$\downarrow$} & 70.6 \textcolor{red}{$\downarrow$} & 49.9 \textcolor{red}{$\downarrow$} & 47.7 \textcolor{red}{$\downarrow$} & 29.5 \textcolor{ForestGreen}{$\uparrow$} & 51.4 \textcolor{ForestGreen}{$\uparrow$} & 30.0 \textcolor{ForestGreen}{$\uparrow$} & \underline{\textbf{66.4}} \textcolor{ForestGreen}{$\uparrow$} & 57.2 \textcolor{red}{$\downarrow$} & 65.6 \textcolor{ForestGreen}{$\uparrow$} & 56.7 \textcolor{red}{$\downarrow$}\\ 
 \rowcolor{gray!20} hFedF & \underline{\textbf{71.5}} \textcolor{ForestGreen}{$\uparrow$} & \underline{\textbf{54.6}} \textcolor{ForestGreen}{$\uparrow$}& \underline{\textbf{75.8}} \textcolor{ForestGreen}{$\uparrow$} & \underline{\textbf{54.9}} \textcolor{ForestGreen}{$\uparrow$} & 49.4 \textcolor{ForestGreen}{$\uparrow$} & \underline{\textbf{29.8}} \textcolor{ForestGreen}{$\uparrow$} & \underline{\textbf{52.1}} \textcolor{ForestGreen}{$\uparrow$} & \underline{\textbf{31.2}} \textcolor{ForestGreen}{$\uparrow$} & 66.1 \textcolor{ForestGreen}{$\uparrow$} & 56.9 \textcolor{red}{$\downarrow$} & \underline{\textbf{66.8}} \textcolor{ForestGreen}{$\uparrow$} & \underline{\textbf{58.1}} \textcolor{ForestGreen}{$\uparrow$} \\
\bottomrule
\end{tabular}
}
\end{center}
\vspace{-15pt}
\end{table}
\endgroup

\begin{figure}[t]
\begin{center}
\centerline{\includegraphics[width=\textwidth]{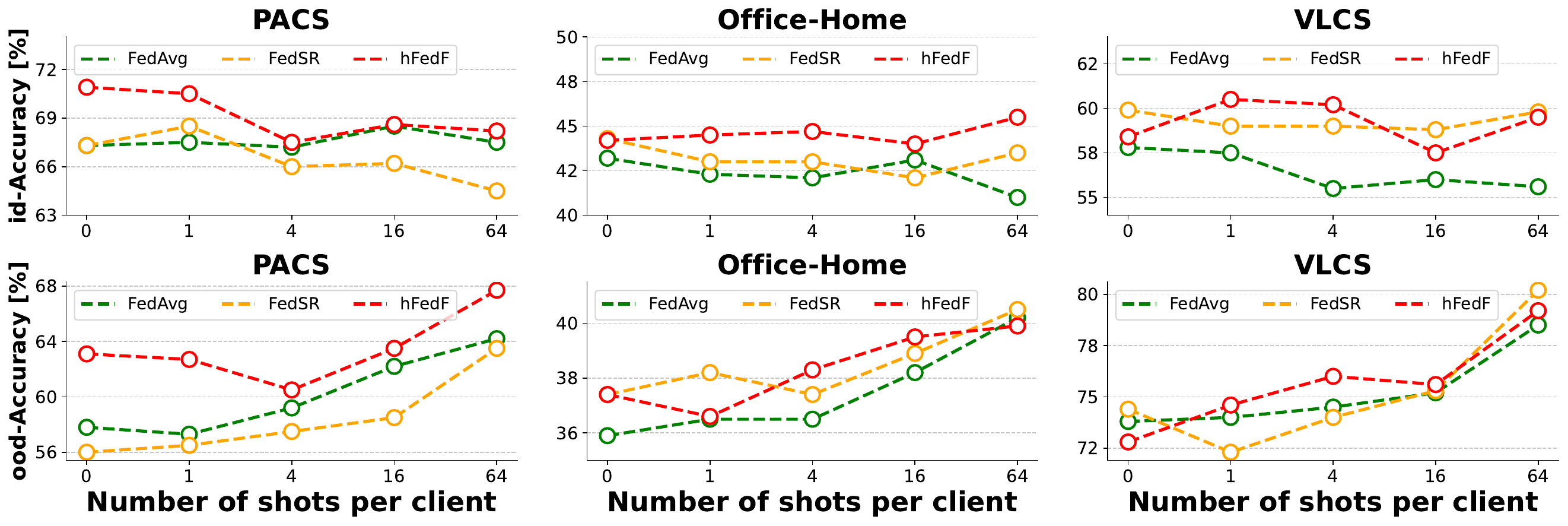}}
\vspace{-5pt}
\caption{Comparison of \textit{id}-accuracy (top row) and \textit{ood}-accuracy (bottom row) across PACS \cite{pacs}, Office-Home \cite{officehome}, and VLCS \cite{vlcs} datasets according to the changes of the number of shots.}
\vspace{-20pt}
\label{fig:nshot}
\end{center}
\end{figure}

\begin{figure}[t]
\includegraphics[width=1.0\linewidth]{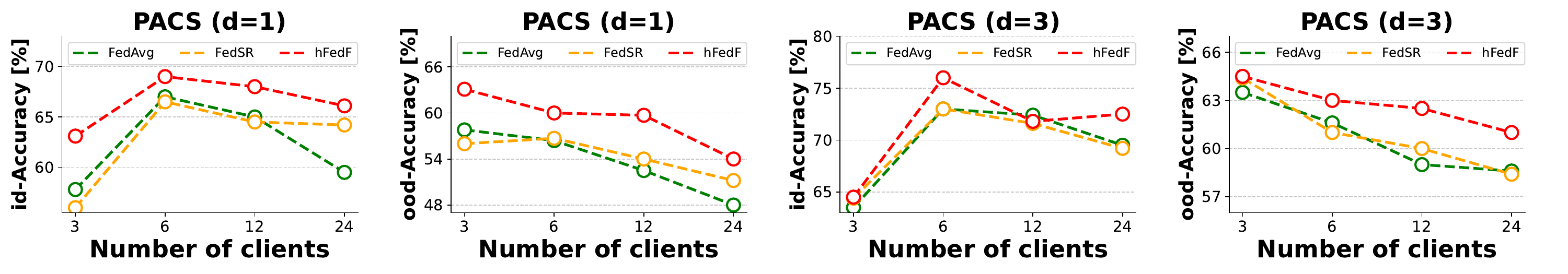}
\caption{Performance comparison of \textit{id}-accuracy (top row) and \textit{ood}-accuracy (bottom row) on PACS dataset with different numbers of clients.}
\vspace{-10pt}
\label{fig:nclients}
\end{figure}


\subsection{Main results} \label{subsec:results}

\textbf{Generalization on domain heterogeneity}\, \autoref{tab:main} shows that our {hFedF} algorithm demonstrates robust improvements in both \textit{id}- and \textit{ood}-accuracy across all datasets compared to other benchmarks. Achieving consistent gains in \textit{id}- and \textit{ood}-accuracy is challenging, especially as most methods experience performance drops when trained for cross-domain generalization. However, {hFedF} maintains superior performance, reflecting its ability to effectively handle domain shifts. In particular, {hFedF} outperforms {pFedHN}, which, despite being hypernetwork-based, shows significantly lower performance. While {FedSR} and {FedGMA} are designed for multi-domain scenarios, they exhibit strengths only in specific cases, such as in parts of the Office-Home dataset, and lack consistent robustness. These methods struggle with the complexity of real-world datasets. In contrast, {hFedF} performs reliably well across PACS, Office-Home, and VLCS, showcasing its resilience and effectiveness in diverse and challenging conditions.

\textbf{From zero- to few-shot learning}\, We also evaluate our approach under a few-shot learning scenario by introducing a small number of target domain samples to each client during training. \Autoref{fig:nshot} illustrates that as the number of shots per client increases, generalization performance (\textit{ood}-accuracy) generally improves across all datasets (PACS, Office-Home, VLCS). This improvement is due to clients gaining specific knowledge about the \textit{ood} domain, which enhances predictions. Simultaneously, the increased data diversity can slightly challenge \textit{id}-accuracy. In most cases, {hFedF} consistently outperforms benchmarks such as FedAvg and the previous state-of-the-art FedSR.

\textbf{Scalability with increasing clients}\, \Autoref{fig:nclients} shows that \textit{hFedF} generally outperforms \textit{FedSR} in \textit{id}- and \textit{ood}-accuracy as the number of clients increases on the PACS dataset for both $d=1$ and $d=3$. This improvement is primarily due to \textit{hFedF}'s ability to handle domain shifts and client heterogeneity effectively through its hypernetwork framework. While \textit{FedSR} remains competitive in certain scenarios, especially with a smaller number of clients, \textit{hFedF} consistently provides scalable and robust performance in varied FL environments, demonstrating its overall advantage.


\begin{figure}[!t]
    \vspace{-5pt}
    \centering
    \begin{subfigure}{0.61\textwidth}
        \includegraphics[width=\linewidth]{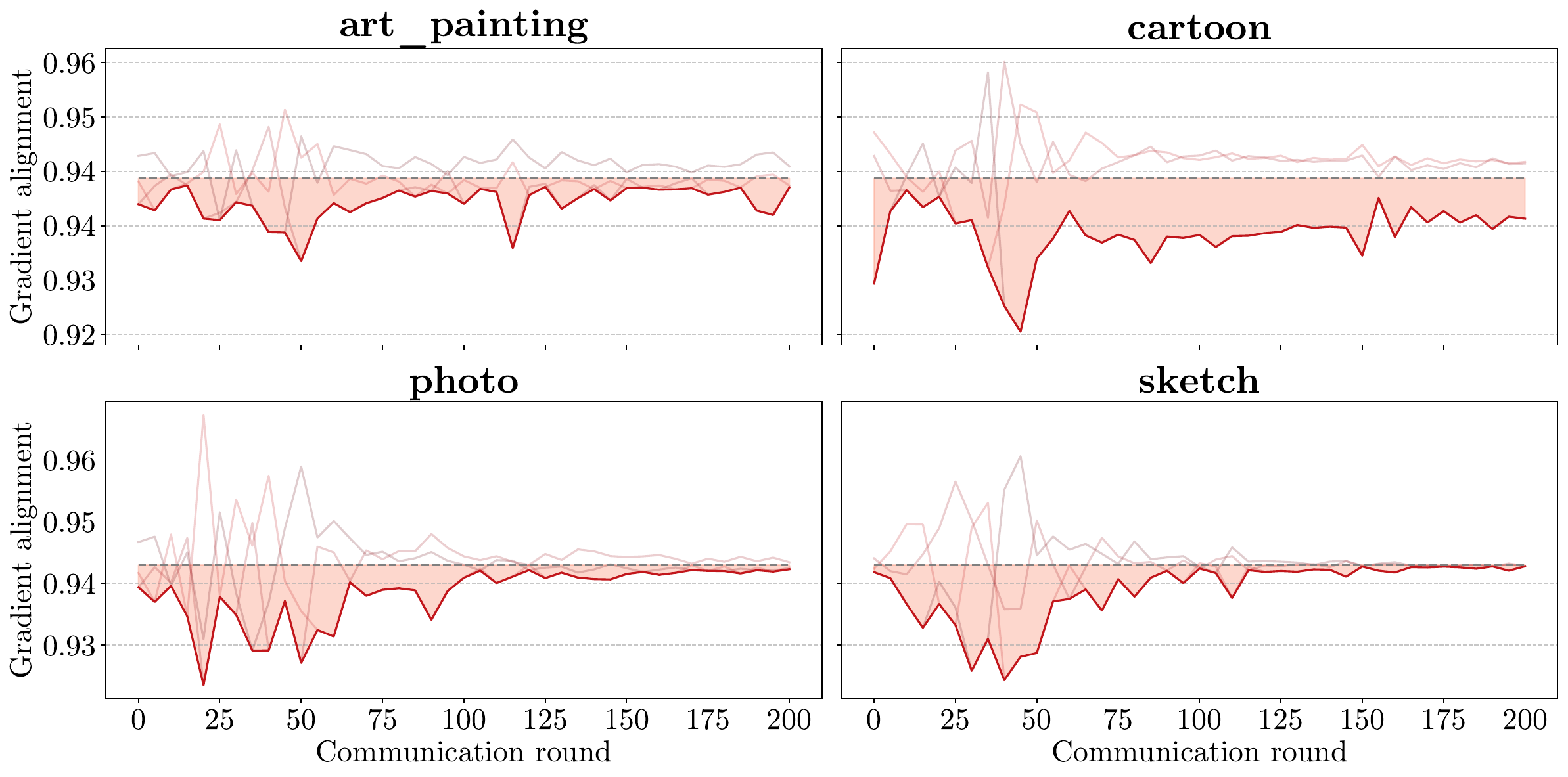}
        \caption{Gradient alignments (i.e., cosine similarity)}     \label{fig:grad_alignment}
    \end{subfigure}
    \begin{subfigure}{0.36\textwidth}
        \includegraphics[width=\linewidth]{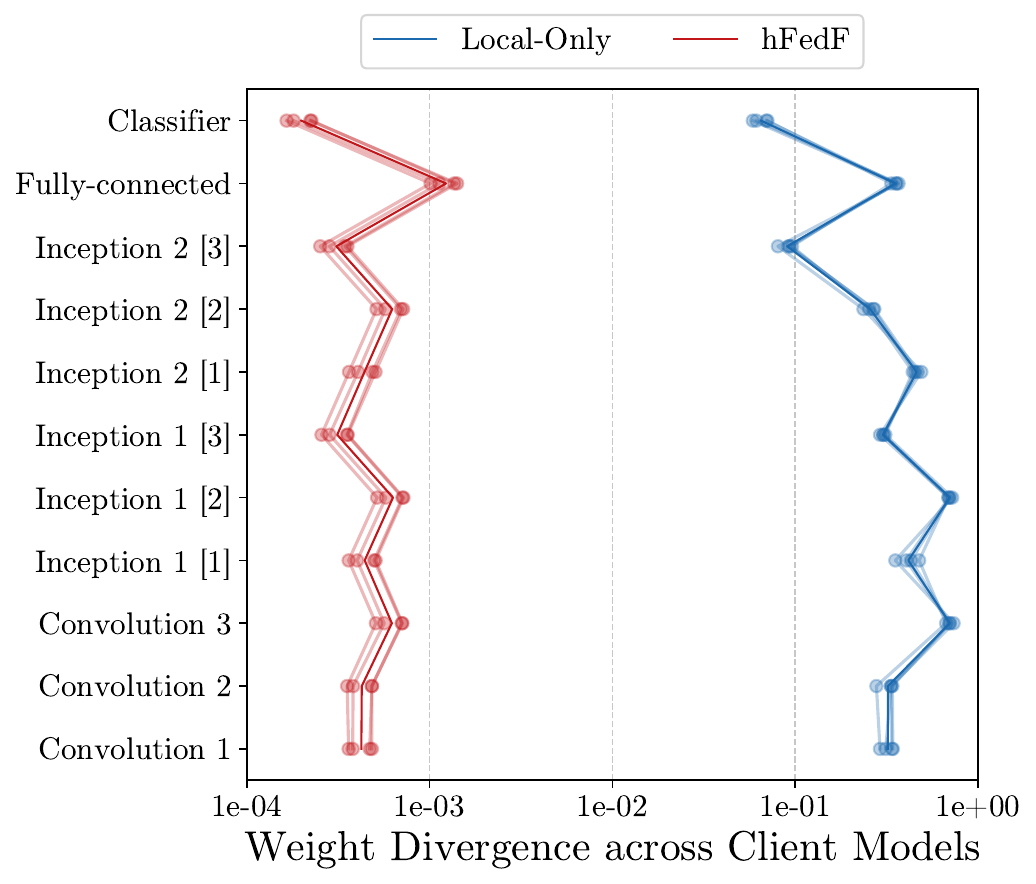}
        \caption{Weight divergence} \label{fig:model_diff}
    \end{subfigure}
    \vspace{-5pt}
    \caption{(a) Convergence of gradient alignment weights, with aggregation weights depicted in transparent color per client over PACS ($d=1$). (b) Euclidean weight divergence\,\cite{fedprox} between client models, showing the weight divergence across all clients, averaged per layer over PACS ($d=1$).}
    \vspace{-10pt}
\end{figure}

\textbf{Analysis on \textsc{GradAlign}}\, \Autoref{fig:grad_alignment} illustrates the behavior and effectiveness of the gradient alignment technique in \textit{hFedF}. The bold lines represent the minimum gradient alignment (i.e., cosine similarity) of each client compared to the average client gradient, depicted by the weight its gradient received during hypernetwork updates. This alignment guides the global update direction, reducing drift over time as weights converge to a common value. Higher weights indicate closer alignment with the consensus direction. Domains like `photo' and `sketch' initially exhibit higher disagreement but achieve better alignment over time compared to `cartoon'. This trend underscores the importance of managing domain interactions to maintain alignment. The analysis reveals that while initial gradient disagreements vary based on domain characteristics, \textsc{GradAlign} appears to help in learning domain-invariant features over time, leading to more stable training across communication rounds.

\textbf{Analysis on weight divergence between local models}\, \Autoref{fig:model_diff} illustrates the Euclidean weight divergence \cite{fedprox} between local client models for our proposed hFedF method compared to locally trained models. The hFedF method exhibits minimal variation across client models, demonstrating its ability to produce a quasi-global model effectively. This reduced variation suggests that the hypernetwork can aggregate knowledge in a manner that maintains consistency across different client models, despite their local updates. The figure shows that, particularly for the deeper layers, the hFedF approach significantly reduces the parameter divergence compared to local training. This indicates that the non-linear fusion of the hypernetwork allows for more stable parameter updates.

\begin{wraptable}{r}{0.5\textwidth}
\vspace{-13pt}
\caption{Averaged accuracy across domains on Office-Home without \textsc{GradAlign} and \textsc{EMA}.}
\label{tab:wo_gradalign_ema}
\resizebox{0.5\textwidth}{!}{%
\begin{tabular}{lllllll} 
\toprule
\multirow{2}{*}{Algorithm} & \multicolumn{2}{c}{$d = 1$} & \multicolumn{2}{c}{$d = 2$} & \multicolumn{2}{c}{$d = 3$} \\ 
\cmidrule{2-7}
  & $\mu_\text{\textit{id}}$ & $\mu_\text{\textit{ood}}$ & $\mu_\text{\textit{id}}$ & $\mu_\text{\textit{ood}}$ & $\mu_\text{\textit{id}}$ & $\mu_\text{\textit{ood}}$    \\ 
\cmidrule{1-7}
hFedF & 49.4 & 29.8 & \underline{\textbf{52.1}} & \underline{\textbf{31.2}} & \underline{\textbf{52.2}} & \underline{\textbf{31.7}}   \\
hFedF w/o \textsc{GradAlign} & 49.2 & \underline{\textbf{30.2}} & 51.9 & 31.2 & 51.5 & 31.4  \\
hFedF w/o \textsc{EMA} & \underline{\textbf{50.9}} & 28.8 & 50.9 & 30.7 & 51.4 & 30.9  \\
hFedF w/o \textsc{GradAlign} \&  \textsc{EMA} & 48.8 & 28.8 & 51.2 & 30.7 & 50.9 & 30.9  \\ 
\bottomrule
\end{tabular}
}
\vspace{-5pt}
\end{wraptable}

\textbf{Ablations on \textsc{EMA} and \textsc{GradAlign}}\, \Autoref{tab:wo_gradalign_ema} presents the ablation study results on the Office-Home dataset, evaluating the impact of excluding \textsc{EMA} and \textsc{GradAlign}. The complete \textit{hFedF} model consistently achieves the highest \textit{id}- and \textit{ood}-accuracy across different domain allocations ($d=1$, $d=2$, $d=3$). Removing \textsc{GradAlign} or \textsc{EMA} results in a slight decrease in performance, indicating their importance in enhancing model stability and alignment. Notably, the combined absence of both techniques further degrades accuracy, underscoring their complementary roles in optimizing federated domain generalization.

\begin{table*}[t]
\caption{Accuracy evaluated on Office-Home \cite{officehome} with $d=3$ ($\mu$: average).}
\vspace{-5pt}
\label{tab:results_officehome_d3}
\resizebox{\textwidth}{!}{
\begin{tabular}{lllllllllllll} 
\toprule
 &  & \multicolumn{5}{c}{$\textit{Acc}_{\text{\textit{id}}}$} &  & \multicolumn{5}{c}{$\textit{Acc}_{\text{\textit{ood}}}$} \\ 
\cmidrule{3-7}\cmidrule{9-13}
 &  & \multicolumn{1}{c}{Art} & \multicolumn{1}{c}{Clipart} & \multicolumn{1}{c}{Product} & \multicolumn{1}{c}{Real World} & \multicolumn{1}{c}{$\mu$} &  & \multicolumn{1}{c}{Art} & \multicolumn{1}{c}{Clipart} & \multicolumn{1}{c}{Product} & \multicolumn{1}{c}{Real World} & \multicolumn{1}{c}{$\mu$} \\ 
\cmidrule{1-1}\cmidrule{3-7}\cmidrule{9-13}
Central &  & 54.1 $\pm$ 0.6 & 47.4 $\pm$ 1.4 & 44.4 $\pm$ 0.7 & 52.1 $\pm$ 1.5 & {49.5} &  & 18.5 $\pm$ 1.1 & 25.7 $\pm$ 0.6 & 35.7 $\pm$ 0.3 & 32.5 $\pm$ 1.1 & {28.1} \\
Local-Only &  & 40.4 $\pm$ 1.9 & 36.4 $\pm$ 3.4 & 32.8 $\pm$ 2.1 & 38.0 $\pm$ 2.4 & {36.9} &  & 13.9 $\pm$ 0.8 & 19.0 $\pm$ 0.8 & 25.5 $\pm$ 0.8 & 24.9 $\pm$ 1.0 & {20.8} \\ 
\cmidrule{1-1}\cmidrule{3-7}\cmidrule{9-13}
FedAvg \cite{fedavg} &  & 56.1 $\pm$ 3.1 & 50.5 $\pm$ 1.5 & 45.9 $\pm$ 3.8 & 54.4 $\pm$ 3.5 & {51.7} &  & 19.5 $\pm$ 0.4 & 26.4 $\pm$ 0.5 & 36.9 $\pm$ 0.4 & 34.8 $\pm$ 0.5 & {29.4} \\
\cmidrule{1-1}\cmidrule{3-7}\cmidrule{9-13}
FedProx \cite{fedprox} &  & 56.1 $\pm$ 2.7 \textcolor{ForestGreen}{$\uparrow$} & 49.3 $\pm$ 2.2 \textcolor{red}{$\downarrow$} & 46.0 $\pm$ 3.6 \textcolor{ForestGreen}{$\uparrow$} & 56.0 $\pm$ 3.1 \textcolor{ForestGreen}{$\uparrow$} & {51.8} \textcolor{ForestGreen}{$\uparrow$} &  & 19.6 $\pm$ 0.2 \textcolor{ForestGreen}{$\uparrow$} & 26.2 $\pm$ 0.2 \textcolor{red}{$\downarrow$} & 38.0 $\pm$ 0.6 \textcolor{ForestGreen}{$\uparrow$} & 35.7 $\pm$ 0.3 \textcolor{ForestGreen}{$\uparrow$} & {29.9} \textcolor{ForestGreen}{$\uparrow$} \\
pFedHN \cite{pfedhn} &  & 48.8 $\pm$ 2.3 \textcolor{red}{$\downarrow$} & 40.5 $\pm$ 6.5 \textcolor{red}{$\downarrow$} & 40.1 $\pm$ 3.5 \textcolor{red}{$\downarrow$} & 46.6 $\pm$ 4.2 \textcolor{red}{$\downarrow$} & {44.0} \textcolor{red}{$\downarrow$} &  & 17.2 $\pm$ 1.0 \textcolor{red}{$\downarrow$} & 22.5 $\pm$ 3.5 \textcolor{red}{$\downarrow$} & 32.4 $\pm$ 2.1 \textcolor{red}{$\downarrow$} & 31.1 $\pm$ 2.5 \textcolor{red}{$\downarrow$} & {25.8} \textcolor{red}{$\downarrow$} \\ 
FedSR \cite{fedsr} &  & 56.3 $\pm$ 2.3 \textcolor{ForestGreen}{$\uparrow$} & \underline{\textbf{50.6 $\pm$ 1.7}} \textcolor{ForestGreen}{$\uparrow$} & 46.3 $\pm$ 2.7 \textcolor{ForestGreen}{$\uparrow$} & \underline{\textbf{54.7 $\pm$ 2.2}} \textcolor{ForestGreen}{$\uparrow$} & {52.0} \textcolor{ForestGreen}{$\uparrow$} &  & 19.7 $\pm$ 1.3 \textcolor{ForestGreen}{$\uparrow$} & 26.6 $\pm$ 0.8 \textcolor{ForestGreen}{$\uparrow$} & 37.9 $\pm$ 0.6 \textcolor{ForestGreen}{$\uparrow$} & 36.4 $\pm$ 0.7 \textcolor{ForestGreen}{$\uparrow$} & {30.2} \textcolor{ForestGreen}{$\uparrow$} \\
FedGMA \cite{fedgma} &  & 56.2 $\pm$ 2.2 \textcolor{ForestGreen}{$\uparrow$} & 49.1 $\pm$ 2.1 \textcolor{red}{$\downarrow$} & 46.3 $\pm$ 3.4 \textcolor{ForestGreen}{$\uparrow$} & 55.4 $\pm$ 3.9 \textcolor{ForestGreen}{$\uparrow$} & {51.7} \textcolor{ForestGreen}{$\uparrow$} &  & 20.3 $\pm$ 0.3 \textcolor{ForestGreen}{$\uparrow$} & 26.3 $\pm$ 0.3 \textcolor{red}{$\downarrow$} & 36.8 $\pm$ 0.7 \textcolor{red}{$\downarrow$} & 35.0 $\pm$ 1.0 \textcolor{ForestGreen}{$\uparrow$} & {29.6} \textcolor{ForestGreen}{$\uparrow$} \\ 
\rowcolor{gray!20} hFedF &  & \underline{\textbf{56.6 $\pm$ 2.3}} \textcolor{ForestGreen}{$\uparrow$} & 50.3 $\pm$ 1.7 \textcolor{red}{$\downarrow$} & \underline{\textbf{47.4 $\pm$ 3.8}} \textcolor{ForestGreen}{$\uparrow$} & 54.4 $\pm$ 2.9 \textcolor{ForestGreen}{$\uparrow$} & \underline{\textbf{52.2}} \textcolor{ForestGreen}{$\uparrow$} &  & \underline{\textbf{20.9 $\pm$ 0.5}} \textcolor{ForestGreen}{$\uparrow$} & \underline{\textbf{29.0 $\pm$ 0.8}} \textcolor{ForestGreen}{$\uparrow$} & \underline{\textbf{39.5 $\pm$ 0.3}} \textcolor{ForestGreen}{$\uparrow$} & \underline{\textbf{37.4 $\pm$ 0.6}} \textcolor{ForestGreen}{$\uparrow$} & \underline{\textbf{31.7}} \textcolor{ForestGreen}{$\uparrow$} \\
\bottomrule
\end{tabular}
}
\end{table*}

\begin{table*}[t]
\caption{Accuracy evaluated on VLCS \cite{vlcs} with $d=3$ ($\mu$: average).}
\vspace{-10pt}
\label{tab:results_vlcs_d3}
\begin{center}
\resizebox{\textwidth}{!}{
\begin{tabular}{lllllllllllll} 
\toprule
 &  & \multicolumn{5}{c}{$\textit{Acc}_{\text{\textit{id}}}$} &  & \multicolumn{5}{c}{$\textit{Acc}_{\text{\textit{ood}}}$} \\ 
\cmidrule{3-7}\cmidrule{9-13}
 &  & \multicolumn{1}{c}{Caltech101} & \multicolumn{1}{c}{LabelMe} & \multicolumn{1}{c}{SUN09} & \multicolumn{1}{c}{PASCAL VOC} & \multicolumn{1}{c}{$\mu$} &  & \multicolumn{1}{c}{Caltech101} & \multicolumn{1}{c}{LabelMe} & \multicolumn{1}{c}{SUN09} & \multicolumn{1}{c}{PASCAL VOC} & \multicolumn{1}{c}{$\mu$} \\ 
\cmidrule{1-1}\cmidrule{3-7}\cmidrule{9-13}
Central &  & 57.0 $\pm$ 2.5 & 65.1 $\pm$ 2.0 & 63.3 $\pm$ 0.8 & 66.0 $\pm$ 1.6 & {62.8} &  & 75.3 $\pm$ 1.9 & 54.1 $\pm$ 0.8 & 48.6 $\pm$ 2.9 & 45.2 $\pm$ 1.1 & {55.8} \\
Local-Only &  & 54.6 $\pm$ 3.2 & 59.4 $\pm$ 4.2 & 61.4 $\pm$ 2.8 & 60.0 $\pm$ 3.7 & {58.8} &  & 63.8 $\pm$ 6.1 & 51.3 $\pm$ 1.5 & 47.6 $\pm$ 2.1 & 43.8 $\pm$ 1.2 & {51.6} \\ 
\cmidrule{1-1}\cmidrule{3-7}\cmidrule{9-13}
FedAvg \cite{fedavg} &  & 61.1 $\pm$ 2.4 & 65.6 $\pm$ 3.8 & 66.5 $\pm$ 2.6 & 67.9 $\pm$ 2.7 & {65.3} &  & 72.8 $\pm$ 0.7 & 54.4 $\pm$ 0.4 & 53.7 $\pm$ 0.5 & 48.7 $\pm$ 1.1 & {57.4} \\
\cmidrule{1-1}\cmidrule{3-7}\cmidrule{9-13}
FedProx \cite{fedprox} &  & 60.2 $\pm$ 2.9 \textcolor{red}{$\downarrow$} & 64.6 $\pm$ 3.0 \textcolor{red}{$\downarrow$} & 67.0 $\pm$ 1.6 \textcolor{ForestGreen}{$\uparrow$} & \underline{\textbf{68.9 $\pm$ 2.7}} \textcolor{ForestGreen}{$\uparrow$} & {65.2} \textcolor{red}{$\downarrow$} &  & 71.6 $\pm$ 0.6 \textcolor{red}{$\downarrow$} & 53.5 $\pm$ 0.8 \textcolor{red}{$\downarrow$} & 51.5 $\pm$ 0.5 \textcolor{red}{$\downarrow$} & 48.0 $\pm$ 0.4 \textcolor{red}{$\downarrow$} & {56.2} \textcolor{red}{$\downarrow$} \\
pFedHN \cite{pfedhn} &  & 59.6 $\pm$ 3.6 \textcolor{red}{$\downarrow$} & 64.8 $\pm$ 3.4 \textcolor{red}{$\downarrow$} & 65.6 $\pm$ 2.8 \textcolor{red}{$\downarrow$} & 67.2 $\pm$ 3.0 \textcolor{red}{$\downarrow$} & {64.3} \textcolor{red}{$\downarrow$} &  & 71.6 $\pm$ 0.9 \textcolor{red}{$\downarrow$} & 55.1 $\pm$ 0.7 \textcolor{ForestGreen}{$\uparrow$} & 51.8 $\pm$ 1.2 \textcolor{red}{$\downarrow$} & 49.5 $\pm$ 1.0 \textcolor{red}{$\downarrow$} & {57.0} \textcolor{red}{$\downarrow$} \\ 
FedSR \cite{fedsr} &  & 60.6 $\pm$ 2.6 \textcolor{red}{$\downarrow$} & 65.3 $\pm$ 2.9 \textcolor{red}{$\downarrow$} & \underline{\textbf{67.8 $\pm$ 3.2}} \textcolor{ForestGreen}{$\uparrow$} & 66.7 $\pm$ 2.7 \textcolor{red}{$\downarrow$} & {65.1} \textcolor{red}{$\downarrow$} &  & 71.7 $\pm$ 2.3 \textcolor{red}{$\downarrow$} & 56.1 $\pm$ 0.7 \textcolor{ForestGreen}{$\uparrow$} & 53.5 $\pm$ 1.1 \textcolor{red}{$\downarrow$} & 48.7 $\pm$ 1.0 \textcolor{ForestGreen}{$\uparrow$} & {57.5} \textcolor{ForestGreen}{$\uparrow$} \\
FedGMA \cite{fedgma} &  & 60.9 $\pm$ 3.6 \textcolor{red}{$\downarrow$} & 65.3 $\pm$ 3.0 \textcolor{red}{$\downarrow$} & 67.0 $\pm$ 2.5 \textcolor{ForestGreen}{$\uparrow$} & 68.2 $\pm$ 3.0 \textcolor{ForestGreen}{$\uparrow$} & {65.3} \textcolor{ForestGreen}{$\uparrow$} &  & 71.9 $\pm$ 0.7 \textcolor{red}{$\downarrow$} & 55.3 $\pm$ 0.5 \textcolor{ForestGreen}{$\uparrow$} & 52.7 $\pm$ 0.2 \textcolor{red}{$\downarrow$} & 48.7 $\pm$ 0.5 \textcolor{ForestGreen}{$\uparrow$} & {57.2} \textcolor{red}{$\downarrow$} \\ 
\rowcolor{gray!20} hFedF &  & \underline{\textbf{61.9 $\pm$ 2.9}} \textcolor{ForestGreen}{$\uparrow$} & \underline{\textbf{66.7 $\pm$ 4.0}} \textcolor{ForestGreen}{$\uparrow$} & 66.2 $\pm$ 2.2 \textcolor{ForestGreen}{$\uparrow$} & 68.1 $\pm$ 2.8 \textcolor{ForestGreen}{$\uparrow$} & \underline{\textbf{65.7}} \textcolor{ForestGreen}{$\uparrow$} &  & \underline{\textbf{75.3 $\pm$ 1.5}} \textcolor{ForestGreen}{$\uparrow$} & \underline{\textbf{55.7 $\pm$ 0.6}} \textcolor{ForestGreen}{$\uparrow$} & \underline{\textbf{53.7 $\pm$ 1.5}} \textcolor{ForestGreen}{$\uparrow$} & \underline{\textbf{50.1 $\pm$ 0.2}} \textcolor{ForestGreen}{$\uparrow$} & \underline{\textbf{58.7}} \textcolor{ForestGreen}{$\uparrow$} \\
\bottomrule
\end{tabular}
}
\end{center}
\vspace{-5pt}
\end{table*}

\textbf{Detailed domain performance on Office-Home and VLCS datasets}\, \Autoref{tab:results_officehome_d3} and \Autoref{tab:results_vlcs_d3} show that {hFedF} consistently achieves the highest \textit{id}- and \textit{ood}-accuracy across both Office-Home and VLCS datasets with $d=3$. These results highlight {hFedF}'s ability to effectively manage domain shifts and its consistent superiority over previous methods in federated domain generalization.

\begin{figure}[!t]
    \vspace{-5pt}
    \centering
    \begin{subfigure}{0.38\textwidth}
        \includegraphics[width=\linewidth]{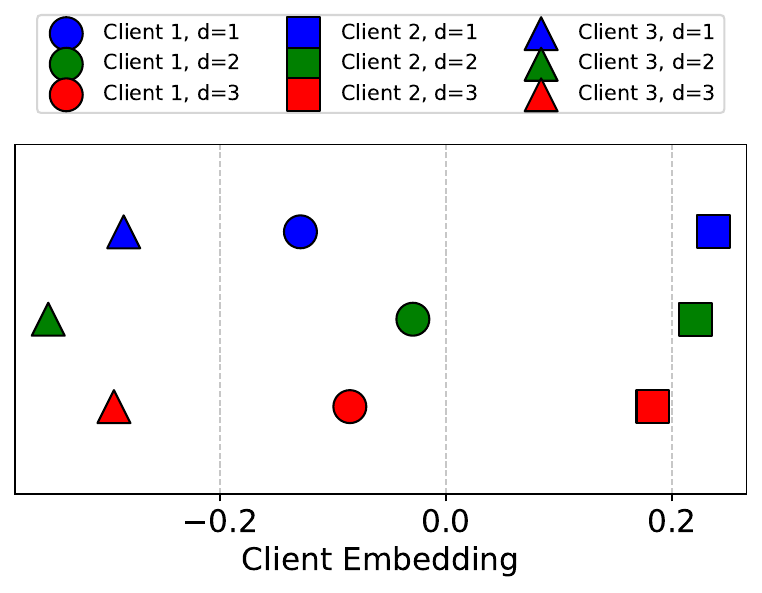}
        \caption{Client embeddings}     \label{fig:embeddings}
    \end{subfigure}
    \begin{subfigure}{0.6\textwidth}
        \includegraphics[width=\linewidth]{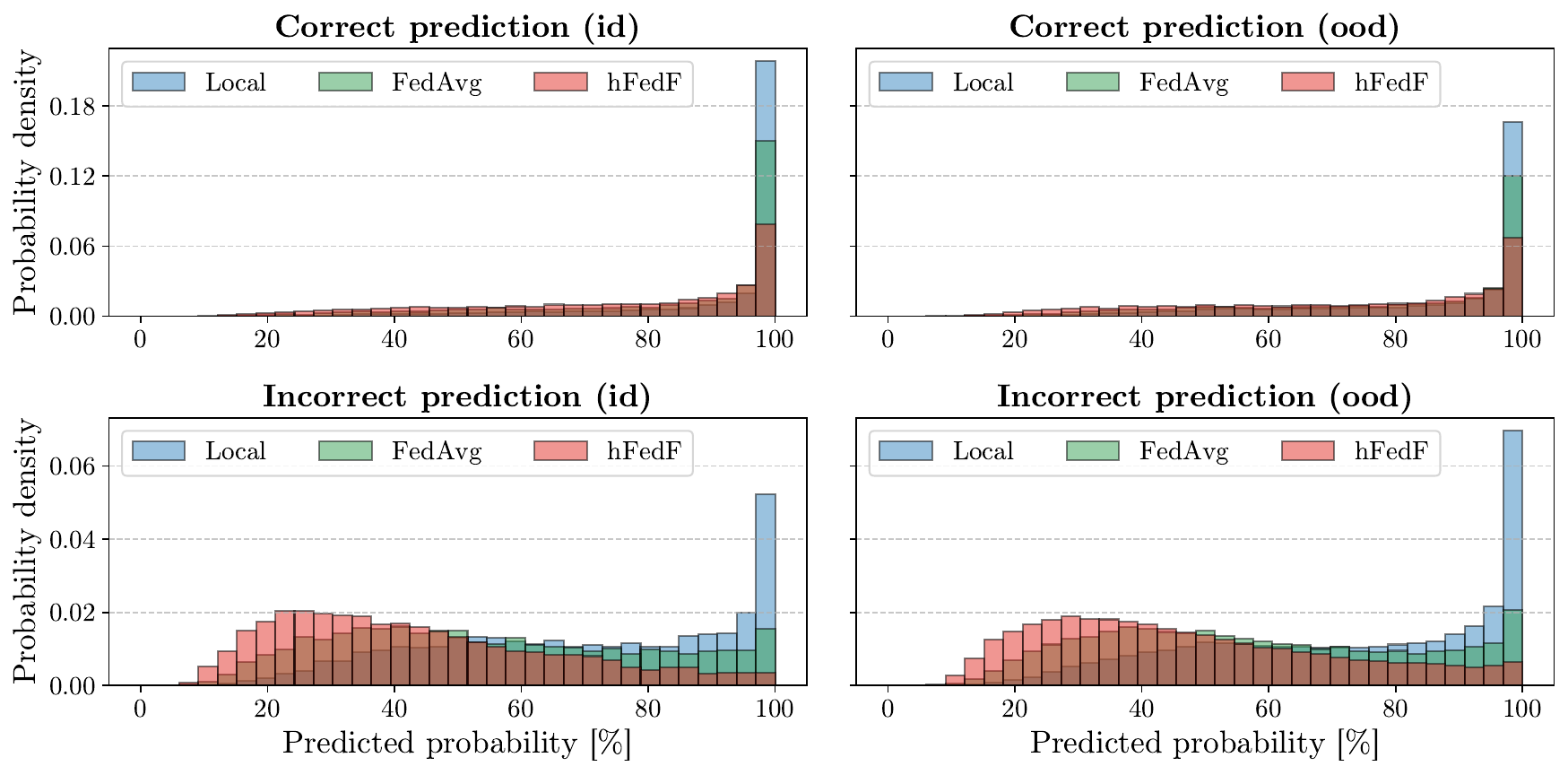}
        \caption{Prediction confidence} \label{fig:bench_pred}
    \end{subfigure}
    \caption{(a) Visualization of 1-dimensional learned client embeddings on Office-Home (Art). The different colors symbolizes different allocation of domains per client and the different marker shape mark the different clients. (b) Histogram of predicted class probabilities for OfficeHome, \(d = 1\)}
    \vspace{-10pt}
\end{figure}


\textbf{Analysis on client embedding}\, \Autoref{fig:embeddings} shows the learned 1-dimensional client embeddings in hFedF for different domain allocations ($d=1$, $d=2$, $d=3$) in the Office-Home dataset. Each subplot represents a specific domain, illustrating how client embeddings vary with different domain allocations. This variability indicates that the client embeddings are influenced by the underlying data distribution, demonstrating the hypernetwork's ability to adapt to diverse data scenarios.


\textbf{Analysis on prediction confidence}\, \Autoref{fig:bench_pred} compares the prediction confidence across different methods: Local, FedAvg, and hFedF. For correct predictions, both in-domain (id) and out-of-domain (ood), hFedF shows a sharper confidence distribution, indicating reliable predictions with high certainty. For incorrect predictions, hFedF exhibits a lower probability density for both id and ood cases, demonstrating a more cautious stance in its predictions. This reduced overconfidence in incorrect predictions suggests that hFedF has a better awareness of its predictive uncertainties. In contrast, FedAvg and Local methods show higher confidence even in their incorrect predictions, indicating a tendency toward overconfidence. The analysis underscores hFedF's ability to effectively learn domain-invariant features, leading to more reliable predictions.
\vspace{-5pt}
\section{Conclusion}
\vspace{-5pt}
This paper introduces \textit{hFedF} (\textbf{h}ypernetwork-based \textbf{Fed}erated \textbf{F}usion), a pioneering approach to Federated Domain Generalization (FDG) that employs hypernetworks for non-linear model aggregation. Traditional domain generalization methods concentrate on learning domain-invariant features; however, the typically linear model averaging in FDG often struggles to manage these effectively. \textit{hFedF} overcomes these obstacles through client-specific embeddings and gradient alignment techniques. Our empirical evaluations across the PACS, Office-Home, and VLCS datasets reveal that \textit{hFedF} consistently achieves superior in-domain and out-of-domain accuracy, occasionally surpassing even centralized training. The scalability of \textit{hFedF} is evidenced by its robust performance across an increasing number of clients, highlighting its adeptness at managing domain-invariant features.
Furthermore, \textit{hFedF} consistently yields lower confidence on incorrect predictions, thereby reducing overconfidence. The approach also exhibits minimal divergence in weights between local client models, indicating effective knowledge aggregation that maintains consistency across varied data distributions. These achievements emphasize the potential of hypernetwork-based methods in advancing FDG and effectively managing domain shifts in federated learning environments.

\clearpage
\bibliographystyle{plainnat}
\bibliography{reference}

\clearpage
\appendix
\section{Broader Impact} \label{sec:appendix} 

Our \textit{hFedF} framework enhances federated learning by effectively handling domain shifts in heterogeneous data while preserving data privacy. The broader impact of our work includes:

\begin{itemize}
    \item \textbf{Enhanced privacy in sensitive domains}: The {hFedF} framework is particularly beneficial in sensitive fields like healthcare and finance, where leveraging diverse, decentralized datasets is crucial without compromising privacy. Our approach addresses domain shifts through client-specific embeddings and non-linear aggregation, ensuring robust performance while maintaining strict privacy standards.

    \item \textbf{Advancement of FDG research}: By tackling the limitations of traditional model aggregation methods and introducing a hypernetwork-based approach, this work advances the field of federated domain generalization. The insights from this study can inspire new strategies for managing domain interactions and improving the stability and effectiveness of federated learning systems. This contribution is significant for the broader field of machine learning, paving the way for future research and development.
    
\end{itemize}

\section{Limitations and Future Work}\label{app:limit}

\subsection{Limitations} 

While \textit{hFedF} advances federated learning by addressing domain shifts, it has certain limitations, particularly with scalability and server computational demands.

\begin{itemize}
    \item \textbf{Scalability with complex models}: The hypernetwork approach in \textit{hFedF} requires predicting numerous parameters for complex models, leading to high memory demands and convergence issues.

    \item \textbf{Computational overhead on the server}: Hypernetwork updates introduce additional computational demands on the server, increasing the overall computational complexity and resource requirements.
\end{itemize}

\subsection{Future Work} 

To address these limitations and enhance \textit{hFedF}, several promising research directions are outlined.

\begin{itemize}
    \item \textbf{Optimizing scalability}: Future work could apply the hypernetwork to a subset of layers or use it for fine-tuning, reducing the number of parameters it needs to predict and alleviating memory and convergence issues.

    \item \textbf{Improving server efficiency}: Research could focus on distributed processing techniques or more efficient update algorithms to reduce the computational burden on the server.

    \item \textbf{Integrating with pretrained models}: Extending \textit{hFedF} to work with large pretrained vision/language models can enhance its scalability and performance without losing the benefits of the hypernetwork-based approach.

    \item \textbf{Adapting to other types of non-iid problems}: Investigating \textit{hFedF}'s application in diverse and complex data distributions will improve its generalization and robustness.
\end{itemize}

\section{Further ablation study}

\subsection{Further analysis on gradient alignment}

\begin{figure}[htb]
\begin{center}
\centerline{\includegraphics[width=\columnwidth]{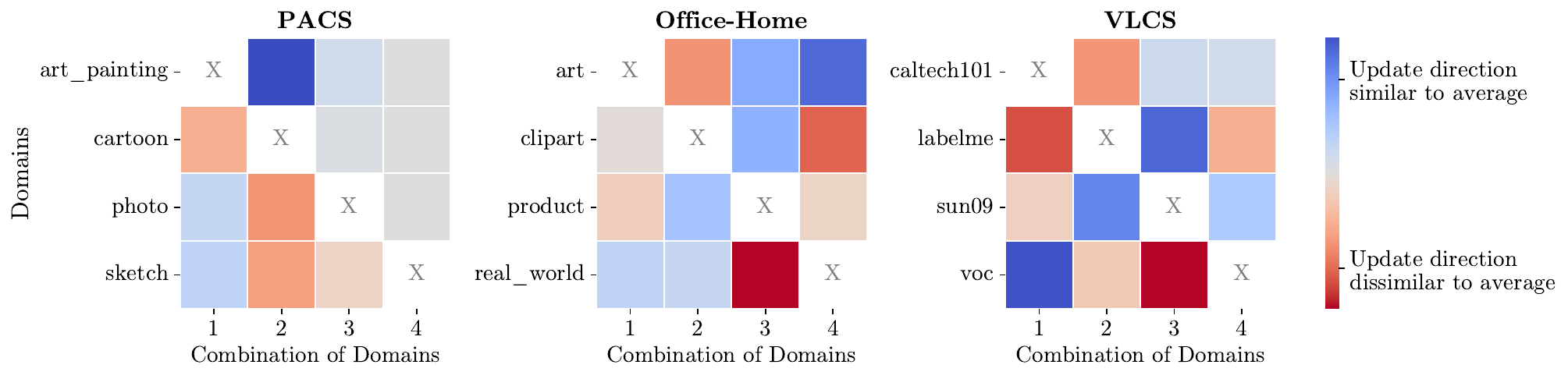}}
\caption{Final gradient alignment weights. At the end of the training (after 200 communication rounds), the aggregation weights of the gradient alignment are visualized (\textit{$d=1$).}}
\label{fig:gradalign_weights}
\end{center}
\vskip -0.2in
\end{figure}

We provide a further examination of the gradient alignment technique used in \textit{hFedF}. \Autoref{fig:gradalign_weights} presents a detailed view of how different domain combinations influence the alignment of update directions with respect to the average gradient across three datasets: PACS, Office-Home, and VLCS. Each subplot shows the gradient alignment weights, with darker shades indicating directions more similar to the average and lighter shades representing more dissimilar directions.

From the figure, we observe that certain domain combinations consistently show higher alignment, while others exhibit significant dissimilarity. For instance, in the PACS dataset, `art\_painting' often aligns closely with the average gradient, whereas `cartoon' shows more variability. In the Office-Home dataset, `art' tends to align better compared to `clipart', which frequently diverges from the average. Similarly, in the VLCS dataset, `sun09' aligns more closely than other categories.

These patterns highlight the importance of managing domain interactions within the federated learning framework. The gradient alignment technique stabilizes training by reducing drift and enhancing the model's ability to learn domain-invariant features. This detailed understanding can inform future research on optimizing FDG algorithms to handle diverse domain combinations effectively.

\subsection{Setting client embeddings from pretrained auto-encoder architecture} \label{appendix:embeddings}
For the main results, client embeddings were learnable with a dimension of 1. To explore alternatives, we used an autoencoder (\Autoref{tab:autoencoder}) to obtain fixed (non-learnable) embeddings based on data distribution. Each client trained a shared autoencoder locally on 100 training images, compressing the images into latent representations. The final embedding was derived by averaging the latent vectors of all validation images, which was then transmitted to the server.

\begin{figure}[h]
\begin{center}
\centerline{\includegraphics[width=\columnwidth]{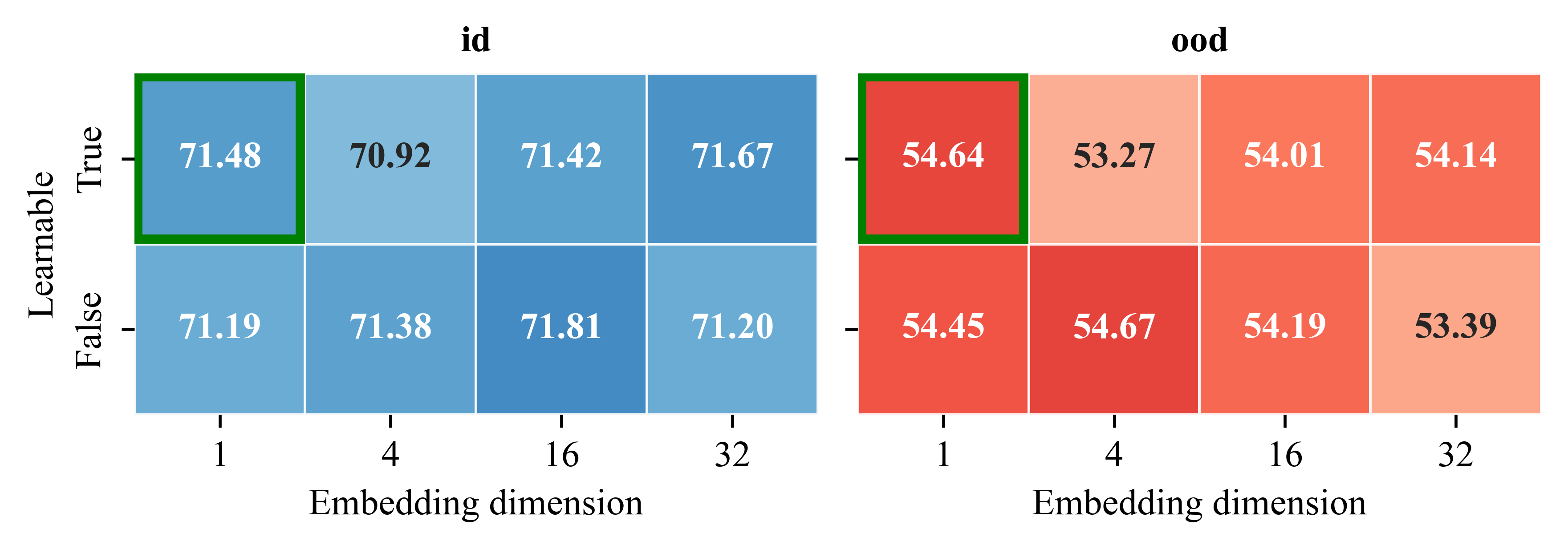}}
\caption{Variations of client embeddings on PACS dataset with $d=1$.}
\label{fig:var_embeddings}
\end{center}
\vskip -0.2in
\end{figure}

\Autoref{fig:var_embeddings} shows performance variations with different embedding dimensions and whether the embeddings are learnable or fixed. The results indicate that these factors do not significantly impact performance. This suggests that our method maintains high performance in both \textit{id} and \textit{ood} scenarios regardless of embedding setup, demonstrating the flexibility and robustness of our approach.

\subsection{Importance of client embedding} \label{appendix:embeddings2}

To clarify the role of client embeddings, we investigated their significance based on empirical evidence. Our analysis is grounded in two main observations: 1) As shown in \Autoref{fig:var_embeddings}, the learnability and dimension of the client embeddings have minimal impact on overall performance, and 2) \Autoref{fig:embeddings} indicates that the hypernetwork prioritizes the overall data distribution over individual client distributions, as disparities between clients are not reflected in the learned embeddings.

To gain further insights, we randomized the client embeddings after each server update and compared the results to the original implementation, as shown in \Autoref{tab:rand_embed}. Within each dataset, the better accuracy is marked in \textbf{bold}. Although hFedF with learned embeddings generally performs better than with randomized embeddings, the differences are not substantial. This finding suggests that the hypernetwork effectively acts as a non-linear fusion of neural networks, largely independent of the specific data distributions of the clients. Additionally, this implies higher data privacy standards since the client embeddings do not contain sensitive information that could be misused. As seen in \Autoref{tab:rand_embed}, even with randomized embeddings, the performance remains robust, supporting the versatility and privacy-preserving nature of our approach.

\paragraph{Another View from random embedding: privacy-secured hFedF} While learning client embeddings has its advantages, a more privacy-secured version of hFedF can be envisioned. By using randomized embeddings, there is no need to send client-specific embeddings to the server. Each client retains its unique key (embedding) to generate the local model without transmitting identifiable information. This method ensures the server remains unaware of client-specific embeddings, providing a higher level of privacy. As shown in \Autoref{tab:rand_embed}, although performance with learned embeddings is generally better, the differences are minimal, indicating that hFedF's effectiveness is largely maintained. Future work could focus on optimizing this approach to balance privacy and performance, potentially leading to a highly personalized and secure hypernetwork framework.

\begin{figure}[t]
\begin{center}
\centerline{\includegraphics[width=\columnwidth]{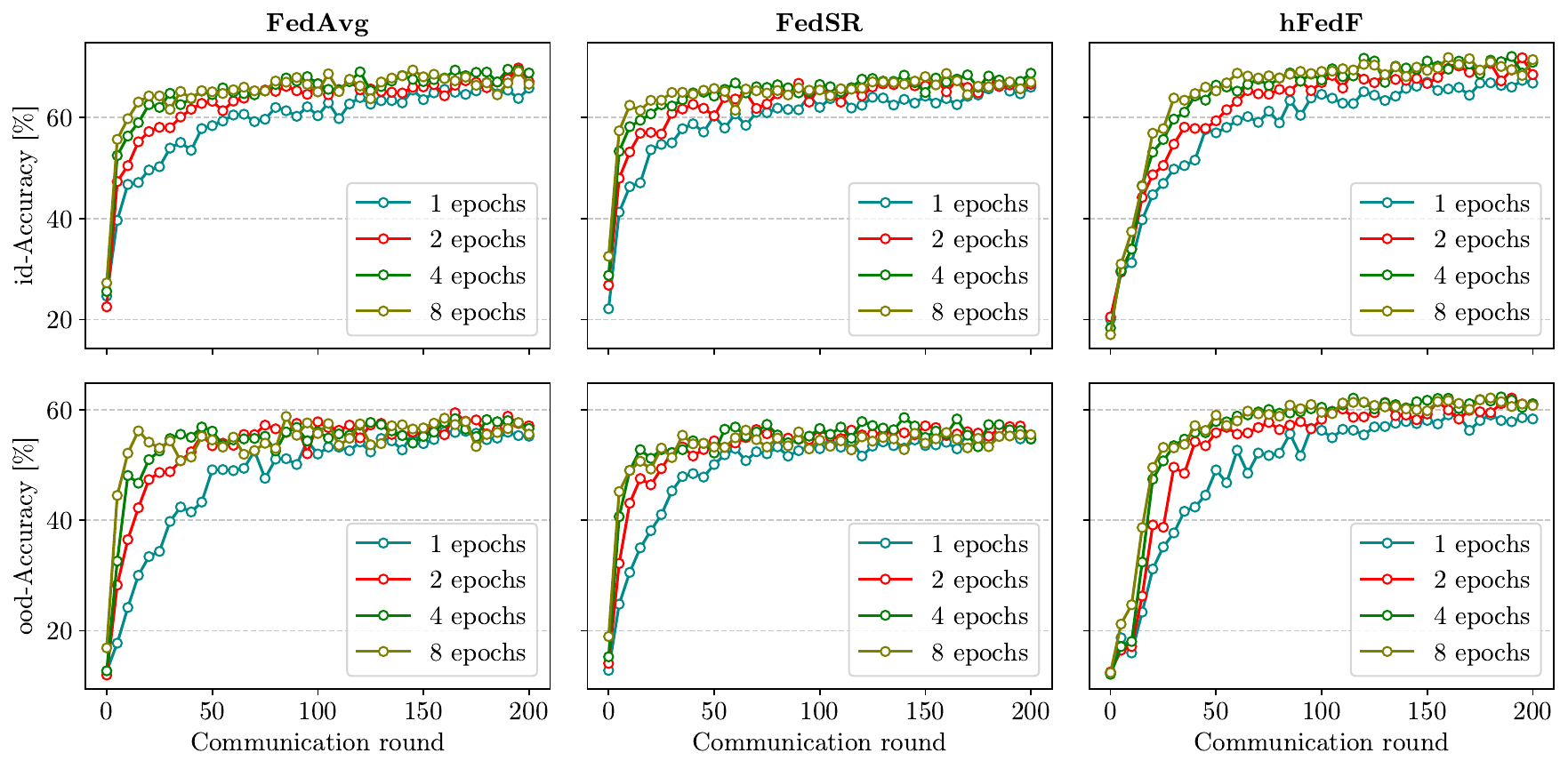}}
\caption{Impact of the number of local epochs on the convergence speed (PACS; "photo" with $d=1$).  }
\label{fig:epochs}
\end{center}
\vskip -0.2in
\end{figure}

\begin{table*}[t]
\caption{Detailed architecture of the Autoencoder.}
\label{tab:autoencoder}
\vskip 0.15in
\begin{center}
\begin{small}
\begin{sc}
\begin{threeparttable}
\begin{tabular}{@{}llll@{}}
\toprule
\textbf{Layer} & \textbf{Output shape}                       & \textbf{Configuration} & \textbf{Activation} \\ \midrule
Conv           & {[}-1, 16, 32, 32{]}                        & k = 3; s = 1; p = 1    & ReLU                \\
MaxPool        & {[}-1, 16, 16, 16{]}                        & k = 2; s = 2; p = 0    & -                   \\
Conv           & {[}-1, 32, 16, 16{]}                        & k = 3; s = 1; p = 1    & ReLU                \\
MaxPool        & {[}-1, 32, 8, 8{]}                          & k = 2; s = 2; p = 0    & -                   \\
Flatten        & {[}-1, 2048{]}                              & -                      & -                   \\
Linear         & {[}-1, $e${]} & -                      & -                   \\
Linear         & {[}-1, 2048{]}                              & -                      & -                   \\
ConvTranspose  & {[}-1, 16, 16, 16{]}                        & k = 2; s = 2; p = 0    & ReLU                \\
ConvTranspose  & {[}-1, 3, 32, 32{]}                         & k = 2; s = 2; p = 0    & Sigmoid  \\
\bottomrule
\end{tabular}
\begin{tablenotes}
\footnotesize
\item[] Abbreviations: Conv - Convolutional; k - kernel size; s - stride; p - padding.
\end{tablenotes}
\end{threeparttable}
\end{sc}
\end{small}
\end{center}
\vskip -0.1in
\end{table*}

\begin{table*}[tb]
\caption{Averaged accuracy across domains on all data sets for randomized client embeddings.}
\label{tab:rand_embed}
\vskip 0.15in
\begin{center}
\begin{small}
\begin{sc}
\begin{tabular}{llllllll} 
\toprule
 \multirow{2}{*}{Data Set} & \multirow{2}{*}{Algorithm} & \multicolumn{2}{c}{$d = 1$} & \multicolumn{2}{c}{$d = 2$} & \multicolumn{2}{c}{$d = 3$} \\ 
\cmidrule{3-8}
 &  & $\mu_\text{\textit{id}}$ & $\mu_\text{\textit{ood}}$ & $\mu_\text{\textit{id}}$ & $\mu_\text{\textit{ood}}$ & $\mu_\text{\textit{id}}$ & $\mu_\text{\textit{ood}}$ \\ 
\midrule
\multirow{2}{*}{\textbf{PACS}} & hFedF & \underline{\textbf{71.5}} & \underline{\textbf{54.6}} & \underline{\textbf{75.8}} & \underline{\textbf{54.9}} & \underline{\textbf{76.1}} & \underline{\textbf{55.2}} \\
 & hFedF (rand. embed.) & 71.1 & 53.9 & 75.1 & 54.8 & 75.7 & 54.5 \\
\midrule
\multirow{2}{*}{\textbf{Office-Home}} & hFedF & \underline{\textbf{49.4}} & 29.8 & \underline{\textbf{52.1}} & 31.2 & \underline{\textbf{52.2}} & \underline{\textbf{31.7}} \\
 & hFedF (rand. embed.) & 48.2 & \underline{\textbf{31.0}} & 51.0 & \underline{\textbf{31.8}} & 51.8 & 30.9 \\ 
\midrule
\multirow{2}{*}{\textbf{VLCS}} & hFedF & \underline{\textbf{66.1}} & \underline{\textbf{56.9}} & \underline{\textbf{66.8}} & \underline{\textbf{58.1}} & \underline{\textbf{65.7}} & \underline{\textbf{58.7}} \\
 & hFedF (rand. embed.) & 65.8 & 55.7 & 66.7 & \underline{\textbf{58.1}} & \underline{\textbf{65.7}} & 58.4 \\ 
\bottomrule
\end{tabular}
\end{sc}
\end{small}
\end{center}
\vskip -0.1in
\end{table*}

\subsection{Choosing the number of local epochs} \label{appendix:moreepochs}
\Autoref{fig:epochs} compares the effects of varying the number of local epochs on convergence speed. As the number of epochs increases, the convergence speed also increases for all algorithms. To balance convergence speed and computational cost, we fixed the number of local epochs to 2 for all experiments.

\subsection{Long-term behavior analyzing with more communication rounds} \label{appendix:morerounds}
\begin{figure}[h]
\begin{center}
\centerline{\includegraphics[width=\columnwidth]{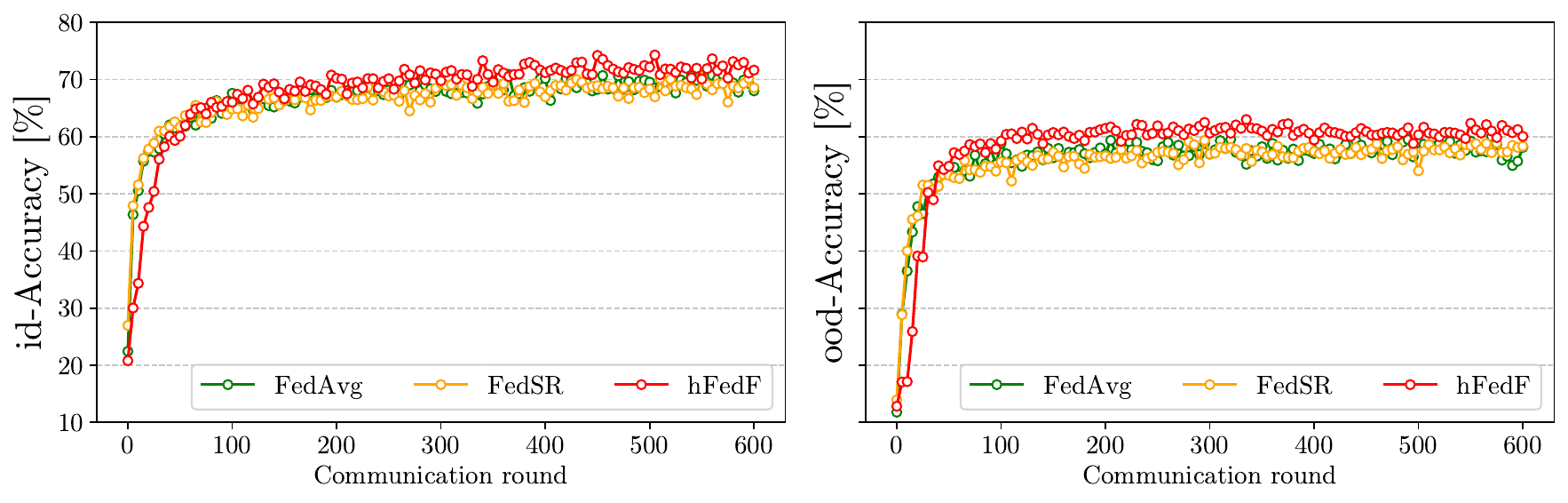}}
\caption{Expanded number of communication rounds (PACS; "photo" with $d=1$).}
\label{fig:rounds}
\end{center}
\vskip -0.2in
\end{figure}

Our hFedF framework demonstrates stable id and ood performance over extended communication rounds, a critical aspect for FL algorithms in life-long learning scenarios. As shown in \Autoref{fig:rounds}, {hFedF} maintains its performance without degradation, and in some cases, even shows an increasing trend in \textit{id} accuracy over time, highlighting its robustness and suitability for long-term FL applications.

\subsection{Impact of scaling EMA} \label{appendix:ema}
\begin{figure}[h]
\begin{center}
\centerline{\includegraphics[width=\columnwidth]{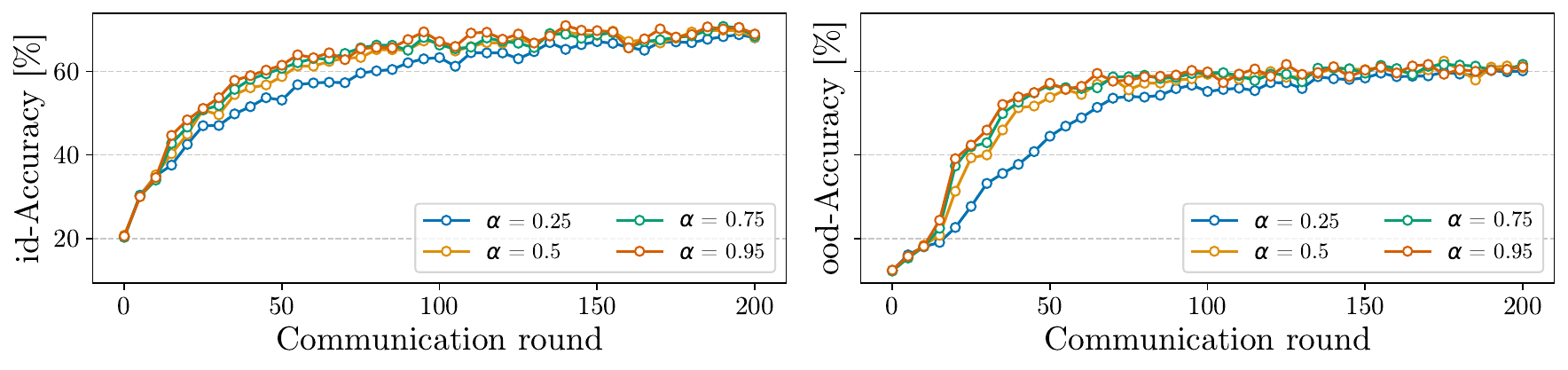}}
\caption{Influence of scaling EMA (PACS; "photo" with $d=1$).}
\label{fig:ema}
\end{center}
\vskip -0.2in
\end{figure}

In \autoref{fig:ema}, the effects of different $\alpha$ values for EMA regularization are visualized. Lower $\alpha$ values result in more regularized hypernetwork updates, leading to flatter but more stable convergence curves. To balance regularization and performance, $\alpha$ values between 0.75 and 0.95 were chosen for the experiments.

\subsection{Local model design for local clients} \label{appendix:models}
The architectures of the hypernetwork (\Autoref{tab:hypernetwork}) and the client model (\Autoref{tab:client_model}) are optimized for layer count, hidden units, activation functions, and convolutional layer configurations. The hypernetwork employs \textit{LeakyReLU} activation functions for effective gradient propagation and the Adam optimizer for improved training stability \cite{hypernetsurvey}. The Auto-encoder's final layer uses a \textit{Sigmoid} function to ensure valid pixel values. The number of clients and the embedding dimension are represented by $N$ and $e$, respectively.

\begin{figure}[htb]
\begin{center}
\centerline{\includegraphics[width=\columnwidth]{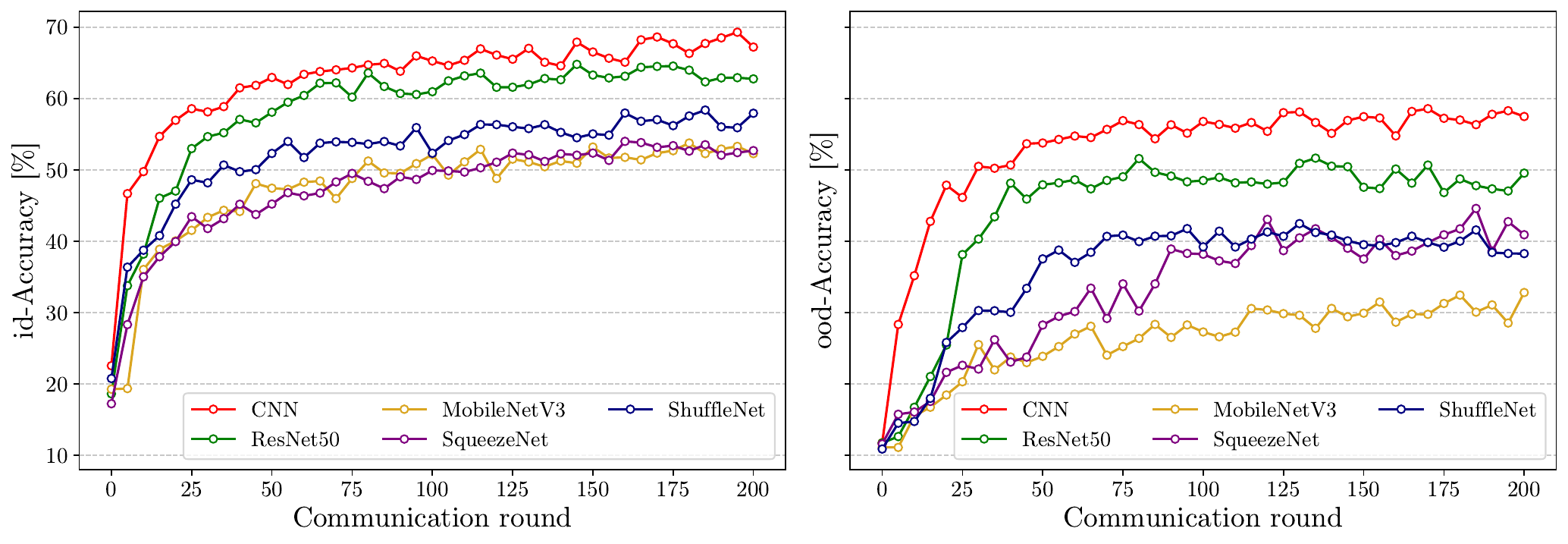}}
\caption{Evaluation of FedAvg on different state-of-the-art architectures as client models (PACS; "photo" with $d=1$).}
\label{fig:nets}
\end{center}
\vskip -0.2in
\end{figure}

Selecting a hypernetwork approach in FL imposes constraints on the client's model architecture. The complexity of the client model directly impacts the hypernetwork's memory demands and its convergence stability. We aimed to design a high-performing client model architecture while avoiding computational complexities and excessive memory consumption. Our exploration of neural architectures focused on minimizing the parameter count while maximizing predictive performance. The resulting architecture is a Convolutional Neural Network (CNN) inspired by the \textit{Inception} architecture \cite{googlenet}, known for its modularity and computational efficiency. To validate our client model design, we compared FedAvg's performance with other established networks, as shown in \Autoref{fig:nets}.

\begin{table*}[t]
\caption{Detailed architecture of the hypernetwork.}
\label{tab:hypernetwork}
\vskip 0.15in
\begin{center}
\begin{small}
\begin{sc}
\begin{threeparttable}
\begin{tabular}{@{}llll@{}}
\toprule
\textbf{Layer}    & \textbf{Output shape}                                             & & \textbf{Activation} \\ \midrule
Embedding\tnote{a}         & {[}1, $e${]} & & -                   \\
Linear\tnote{b}            & {[}1, 50{]}                                                      & & LeakyReLU           \\
Linear\tnote{b}            & {[}1, 50{]}                                                      & & LeakyReLU           \\
Linear\tnote{b}            & {[}1, 50{]}                                                      & & LeakyReLU           \\
Linear\tnote{b}            & {[}1, 50{]}                                                      & & -                   \\
Multi-Head Linear\tnote{b} & {[}1, $n_{\text{layers}}$, $n_{\text{parameters of layer}}${]}    & & -                  \\
\bottomrule
\end{tabular}
\begin{tablenotes}
\footnotesize
\item[a] The parameters of this layer are defined by $\nu$. The embedding layer $\nu$ has a shape of $[N, e]$. The hypernetwork takes the client ID as an index and uses it to retrieve the corresponding embedding from the embedding layer.
\item[b] The parameters of these layers are defined by $\theta$.
\end{tablenotes}
\end{threeparttable}
\end{sc}
\end{small}
\end{center}
\vskip -0.1in
\end{table*}

\begin{table*}[t]
\caption{Detailed architecture of the client model.}
\label{tab:client_model}
\vskip 0.15in
\begin{center}
\begin{small}
\begin{sc}
\begin{threeparttable}
\begin{tabular}{@{}llll@{}}
\toprule
\textbf{Layer}     & \textbf{Output shape}          & \textbf{Configuration}           & \textbf{Activation} \\ \midrule
Convolutional      & {[}-1, 32, 32, 32{]}           & k = 3; s = 1; p = 1              & -                   \\
MaxPool         & {[}-1, 32, 16, 16{]}           & k = 2; s = 2; p = 0              & ReLU                \\
Convolutional     & {[}-1, 32, 16, 16{]}           & k = 1; s = 1; p = 0              & -                   \\
Convolutional      & {[}-1, 64, 16, 16{]}           & k = 3; s = 1; p = 1              & -                   \\
MaxPool         & {[}-1, 64, 8, 8{]}             & k = 2; s = 2; p = 0              & ReLU                \\
Inception module   & {[}-1, 176, 8, 8{]}            & int. channels = {[}32, 64, 16{]} & ReLU                \\
Inception module   & {[}-1, 288, 8, 8{]}            & int. channels = {[}32, 64, 16{]} & ReLU                \\
AdaptiveAvgPool & {[}-1, 288, 3, 3{]}            & output size = 3                  & -                   \\
Flatten            & {[}-1, 2592{]}                 & -                                & -                   \\
Dropout            & {[}-1, 2592{]}                 & ratio = 0.2                      & -                   \\
Linear             & {[}-1, 256{]}                  & -                                & -                   \\
Linear             & {[}-1, $n_{\text{classes}}${]} & -                                & -                   \\ 
\bottomrule
\end{tabular}
\begin{tablenotes}
\footnotesize
\item[] Abbreviations: k - kernel size; s - stride; p - padding; int. - intermediate.
\end{tablenotes}
\end{threeparttable}
\end{sc}
\end{small}
\end{center}
\vskip -0.1in
\end{table*}

\subsection{Experimental Details} \label{appendix:experiments}
\subsubsection{Algorithms} \label{appendix:algos}

Our proposed strategy for allocating source domains to clients effectively manages the severity of domain shifts. The algorithm begins by calculating the minimum number of partitions \(a\) required for each domain. It also determines the number of domains \(b\) that need to be split into one additional partition. This step is essential because not all domains can be uniformly split to meet the requirement of having exactly \(d\) domains per client. The domains are then divided evenly, with the \(b\) largest domains being split into an extra partition if necessary. Finally, the partitions are randomly distributed among the clients, ensuring that each client receives exactly \(d\) domains. This method maintains a balanced allocation of domain data, effectively managing domain shifts across clients. For the sake of completeness and reproducibility, the supplementary algorithms (\Autoref{algo:clientupdate}, \Autoref{algo:ema}, \Autoref{algo:datasplit}) to \Autoref{algo:fedtrain} are provided.

\begin{algorithm}[hbt]
\begin{algorithmic}
    \caption{Client Update (\textbf{ClientUpdate})} \label{algo:clientupdate}
    \STATE {\bfseries Require:} $\mu$ - learning rate of client, $\mathcal{D}$ - data of client, $f(\cdot; \tilde\varphi)$ - client model
    \FOR{x, y $\in \mathcal{D}$}
        \STATE $\hat{y} \gets f_c(x; \tilde\varphi)$
        \STATE $\mathcal{L} \gets \ell_{\text{cross-entropy}}(y, \hat{y})$
        \STATE $\tilde\varphi \gets \tilde\varphi - \mu \cdot \nabla_{\varphi} \mathcal{L}$
    \ENDFOR
    \STATE {\bfseries Return $i$:} $\tilde\varphi$
\end{algorithmic}
\end{algorithm}

\begin{algorithm}[hbt]
\begin{algorithmic}
    \caption{Gradient Alignment (\textbf{GradAlign})} \label{algo:gradalign}
    \STATE {\bfseries Require:} $\mathcal{S}$ - set of clients, $g_i$ - local gradient of client $i$
    \STATE compute $g_{\text{avg}} \gets \frac{1}{|\mathcal{S}|}\sum_{i \in \mathcal{S}} g_i$
    \FOR{client $i \in \mathcal{S}$}
        \STATE $\gamma_i^t \gets \textit{cos}\ (g_{\text{avg}}, g_i)$
        \STATE $\tilde{\gamma}_i \gets \textit{softmax}\ (\gamma_i)$
    \ENDFOR
    \STATE {\bfseries Return (client $i$):} $\tilde{\gamma}_i$
\end{algorithmic}
\end{algorithm}

\begin{algorithm}[hbt]
\begin{algorithmic}
    \caption{Exponential Moving Average (\textbf{EMA})} \label{algo:ema}
    \STATE {\bfseries Require:} $\theta^t$ - parameters of hypernetwork, $t$ - communication round, $w \in \mathbb{Z}^+$ - warm-up round, $\alpha \in [0, 1]$ - smoothing factor
    \IF{warm-up round is reached ($t = w$)}
    \STATE {\bfseries Initialize:} $\theta_{\text{EMA}}^t \gets \theta^t$
    \ENDIF
    \IF{$t > w$}
        \STATE $\theta_{\text{EMA}}^{t+1} \gets \alpha \theta^t + (1-\alpha) \theta_{\text{EMA}}^t$
        \STATE $\theta^t \gets \theta_{\text{EMA}}^{t+1}$
    \ENDIF
    \STATE {\bfseries Return:} $\theta^t$
\end{algorithmic}
\end{algorithm}

\begin{algorithm}[hbt]
\begin{algorithmic}
    \caption{Data split} \label{algo:datasplit}
    \STATE {\bfseries Require:} $N > 1$ - number of clients, $d$ - number of source domains per client, $\{\mathcal{D}_j\}_{j \in \mathcal{Z}_{\text{src}}}$ - data sets of source domains

    \STATE $a \gets \big\lfloor \frac{N \cdot d}{|\mathcal{Z}_{\text{src}}|} \big\rfloor$
    \STATE $b \gets \textit{mod }(N \cdot d, |\mathcal{Z}_{\text{src}}|)$
    \FOR{ {\bfseries each} $\mathcal{D}_j$}
        \IF{$\mathcal{D}_j$ is part of $b$ largest data sets}
            \STATE $\textit{SplitRule} \gets [\frac{1}{a+1},..., \frac{1}{a+1}]^{a+1}$
        \ELSE
            \STATE $\textit{SplitRule} \gets [\frac{1}{a},..., \frac{1}{a}]^a$
        \ENDIF
        \STATE split $\mathcal{D}_j$ randomly according to \textit{SplitRule}
        \STATE $\text{subsets}_j \gets$ split $\mathcal{D}_j$
    \ENDFOR
    \FOR{1 {\bfseries to} d}
        \FOR{client $i \in \{1, ..., N\}$}
            \FOR{domain $j \in \mathcal{Z}_{\text{src}}$}
                \IF{$\text{subsets}_j$ has splits}
                    \STATE append first element of $\text{subsets}_j$ to $\text{subsets}_i$
                    \STATE remove first element of $\text{subsets}_j$
                    \STATE break out of inner loop
                \ENDIF
            \ENDFOR
        \ENDFOR
    \ENDFOR
    \FOR{client $i \in \{1, ..., N\}$}
        \STATE concatenate elements in $\text{subsets}_i$
        \STATE $\mathcal{D}_i \gets$ concatenated $\text{subsets}_i$
    \ENDFOR
    \STATE {\bfseries Return (client $i$):} $\mathcal{D}_i$
\end{algorithmic}
\end{algorithm}

\clearpage

\subsubsection{Hyperparameter search} \label{appendix:hyperparams}
\Autoref{tab:hyperparameters} depicts the search grid and the selected values for each hyperparameter, which was tuned to benchmark the algorithm. Thereby, the goal is to maintain fairness in the tuning process without overextending computational resources. For the non-federated methods, namely \textit{Central} and \textit{Local}, the hyperparameters of FedAvg are adopted. A consistent search grid was chosen across all data sets and algorithms, where the reference values are determined either by consulting the original papers of the respective algorithms or by adhering to the benchmarking suggestions for FDG \cite{feddg_benchmarks}. The final value is chosen based on the optimal performance observed in a single run with a specific seed for a designated target domain after 100 communication rounds.

\begin{table*}[htb]\scriptsize
\caption{Detailed search grid and selected values of hyperparameters for all algorithms and data sets.}
\label{tab:hyperparameters}
\vskip 0.15in
\begin{center}
\begin{sc}
\begin{tabular}{lllccccc} 
\toprule
\multirow{2}{*}{\textbf{Algorithm}} & \multirow{2}{*}{\textbf{Hyperparameter}} &  & \multirow{2}{*}{\textbf{Grid }} & \multicolumn{1}{l}{} & \multicolumn{3}{c}{\textbf{Selected Value}} \\ 
\cmidrule{6-8}
 &  &  &  & \multicolumn{1}{l}{} & PACS & Office-Home & VLCS \\ 
\cmidrule{1-2}\cmidrule{4-4}\cmidrule{6-8}
\multirow{2}{*}{FedAvg} & Client learning rate &  & \{1e-3, 1e-4, 1e-5\} &  & 1e-3 & 1e-3 & 1e-4 \\
 & Client weight decay &  & \{1e-3, 1e-4, 1e-5\} &  & 1e-4 & 1e-5 & 1e-4 \\ 
\cmidrule{1-2}\cmidrule{4-4}\cmidrule{6-8}
\multirow{3}{*}{FedProx} & Client learning rate &  & \{1e-3, 1e-4, 1e-5\} &  & 1e-3 & 1e-3 & 1e-3 \\
 & Client weight decay &  & \{1e-3, 1e-4, 1e-5\} &  & 1e-4 & 1e-4 & 1e-4 \\
 & $L_2$ regularizer &  & \{1e-1, 1e-2, 1e-3\} &  & 1e-2 & 1e-2 & 1e-2 \\ 
\cmidrule{1-2}\cmidrule{4-4}\cmidrule{6-8}
\multirow{4}{*}{pFedHN} & Server learning rate &  & \{1e-1, 1e-2, 1e-3\} &  & 1e-2 & 1e-3 & 1e-3 \\
 & Server weight decay &  & \{1e-3, 1e-4, 1e-5\} &  & 1e-3 & 1e-4 & 1e-4 \\
 & Client learning rate &  & \{1e-3, 1e-4, 1e-5\} &  & 1e-3 & 1e-3 & 1e-3 \\
 & Client weight decay &  & \{1e-3, 1e-4, 1e-5\} &  & 1e-5 & 1e-4 & 1e-4 \\ 
\cmidrule{1-2}\cmidrule{4-4}\cmidrule{6-8}
\multirow{3}{*}{FedSR} & Client learning rate &  & \{1e-3, 1e-4, 1e-5\} &  & 1e-3 & 1e-3 & 1e-3 \\
 & Client weight decay &  & \{1e-3, 1e-4, 1e-5\} &  & 1e-5 & 1e-4 & 1e-4 \\
 & $L_2$ regularizer &  & \{1e-1, 1e-2, 1e-3\} &  & 1e-1 & 1e-1 & 1e-2 \\ 
\cmidrule{1-2}\cmidrule{4-4}\cmidrule{6-8}
\multirow{4}{*}{FedGMA} & Client learning rate &  & \{1e-3, 1e-4, 1e-5\} &  & 1e-3 & 1e-3 & 1e-3 \\
 & Client weight decay &  & \{1e-3, 1e-4, 1e-5\} &  & 1e-4 & 1e-4 & 1e-4 \\
 & Mask threshold &  & \{0.3, 0.5, 0.7\} &  & 0.7 & 0.3 & 0.5 \\
 & Server step size &  & \{1e-0, 1e-1, 1e-2\} &  & 1e-0 & 1e-2 & 1e-2 \\ 
\cmidrule{1-2}\cmidrule{4-4}\cmidrule{6-8}
\multirow{5}{*}{hFedF} & Server learning rate &  & \{1e-1, 1e-2, 1e-3\} &  & 1e-3 & 1e-3 & 1e-3 \\
 & Server weight decay &  & \{1e-3, 1e-4, 1e-5\} &  & 1e-5 & 1e-5 & 1e-3 \\
 & Server EMA decay &  & \{0.75, 0.85, 0.95\} &  & 0.95 & 0.75 & 0.75 \\
 & Client learning rate &  & \{1e-3, 1e-4, 1e-5\} &  & 1e-3 & 1e-3 & 1e-3 \\
 & Client weight decay &  & \{1e-3, 1e-4, 1e-5\} &  & 1e-3 & 1e-3 & 1e-3 \\
\bottomrule
\end{tabular}
\end{sc}
\end{center}
\vskip -0.1in
\end{table*}

\clearpage

\onecolumn
\clearpage
\vfill
\subsection{Detailed Results} \label{appendix:results}
The detailed results are derived from the final evaluation after 200 communication rounds, averaged across three seeds - including the standard deviation. The \textit{id}- and \textit{ood}-accuracy are separately stated for each left-out target domain. Each data set has in total 4 domains, where 3 are seen overall during training, and 1 is left-out for \textit{ood}-testing. The averages across all combinations of source and target domain are marked in \textbf{bold} and correspond to the same numbers reported in \Autoref{subsec:results}. Since the centralized method holds in every case 3 source domains, its performances are only presented for $d=3$. The training curves are averaged across target domains and seeds, and are measured every 5 communication rounds. All computations are conducted on two NVIDIA RTX A5000 24GB GPUs.

\subsubsection{Performance on PACS dataset}\label{appendix:pacs}
Our comprehensive results on PACS dataset detailed in \Autoref{tab:results_pacs_d1}, \Autoref{tab:results_pacs_d2}, and \Autoref{tab:results_pacs_d3} demonstrate {hFedF}'s superior domain generalization performance, consistently outperforming benchmarks across all domain complexities. 

\begin{table*}[htb]
\caption{Accuracy evaluated on PACS; $d=1$.}
\label{tab:results_pacs_d1}
\vskip 0.15in
\begin{center}
\begin{small}
\begin{sc}
\resizebox{0.95\textwidth}{!}{
\begin{threeparttable}
\begin{tabular}{lllllllllllll} 
\toprule
 &  & \multicolumn{5}{c}{$\textit{Acc}_{\text{\textit{id}}}$} &  & \multicolumn{5}{c}{$\textit{Acc}_{\text{\textit{ood}}}$} \\ 
\cmidrule{3-7}\cmidrule{9-13}
 &  & \multicolumn{1}{c}{A} & \multicolumn{1}{c}{C} & \multicolumn{1}{c}{P} & \multicolumn{1}{c}{S} & \multicolumn{1}{c}{$\mu$} &  & \multicolumn{1}{c}{A} & \multicolumn{1}{c}{C} & \multicolumn{1}{c}{P} & \multicolumn{1}{c}{S} & \multicolumn{1}{c}{$\mu$} \\ 
\cmidrule{1-1}\cmidrule{3-7}\cmidrule{9-13}
Central &  & - & - & - & - & \textbf{-} &  & - & - & - & - & \textbf{-} \\
Local &  & 82.3 $\pm$ 3.6 & 74.9 $\pm$ 12.6 & 74.4 $\pm$ 12.1 & 73.8 $\pm$ 9.7 & \textbf{76.4} &  & 28.4 $\pm$ 8.0 & 34.3 $\pm$ 6.5 & 41.9 $\pm$ 18.5 & 38.8 $\pm$ 6.4 & \textbf{35.8} \\ 
\cmidrule{1-1}\cmidrule{3-7}\cmidrule{9-13}
FedAvg &  & 73.4 $\pm$ 9.5 & 69.8 $\pm$ 15.3 & 67.3 $\pm$ 12.8 & 70.0 $\pm$ 6.1 & \textbf{70.1} &  & 37.8 $\pm$ 1.1 & 50.4 $\pm$ 0.6 & 57.8 $\pm$ 1.0 & 59.2 $\pm$ 1.5 & \textbf{51.3} \\
FedProx &  & 73.8 $\pm$ 8.8 & 70.0 $\pm$ 13.6 & 67.2 $\pm$ 14.0 & 68.7 $\pm$ 8.2 & \textbf{69.9} &  & 36.1 $\pm$ 1.7 & 51.7 $\pm$ 1.8 & 56.1 $\pm$ 0.5 & 58.7 $\pm$ 1.3 & \textbf{50.7} \\
pFedHN &  & 69.8 $\pm$ 9.9 & 65.2 $\pm$ 14.1 & 60.3 $\pm$ 11.1 & 65.4 $\pm$ 10.8 & \textbf{65.2} &  & 36.0 $\pm$ 6.9 & 47.1 $\pm$ 4.1 & 55.7 $\pm$ 6.1 & 54.2 $\pm$ 1.3 & \textbf{48.2} \\ 
\cmidrule{1-1}\cmidrule{3-7}\cmidrule{9-13}
FedSR &  & 73.0 $\pm$ 9.3 & 67.5 $\pm$ 14.2 & 67.3 $\pm$ 13.7 & 68.9 $\pm$ 9.4 & \textbf{69.2} &  & 35.8 $\pm$ 0.7 & 47.7 $\pm$ 2.3 & 56.0 $\pm$ 1.0 & 55.8 $\pm$ 0.8 & \textbf{48.9} \\
FedGMA &  & 73.6 $\pm$ 8.4 & 67.8 $\pm$ 14.0 & 65.3 $\pm$ 14.5 & 67.9 $\pm$ 9.1 & \textbf{68.7} &  & 37.4 $\pm$ 1.0 & 49.8 $\pm$ 2.3 & 56.3 $\pm$ 2.0 & 54.8 $\pm$ 1.0 & \textbf{49.6} \\ 
\cmidrule{1-1}\cmidrule{3-7}\cmidrule{9-13}
hFedF &  & 75.4 $\pm$ 6.1 & 70.6 $\pm$ 14.0 & 70.9 $\pm$ 7.5 & 69.1 $\pm$ 10.6 & \textbf{71.5} &  & 42.4 $\pm$ 0.3 & 51.9 $\pm$ 0.3 & 63.1 $\pm$ 1.3 & 61.2 $\pm$ 0.9 & \textbf{54.6} \\
\bottomrule
\end{tabular}
\begin{tablenotes}
\footnotesize
\item[] Abbreviations: $\mu$ - average; A - art painting; C - cartoon; P - photo; S - sketch.
\end{tablenotes}
\end{threeparttable}
}
\end{sc}
\end{small}
\end{center}
\vskip -0.1in
\end{table*}

\begin{table*}[htb]
\caption{Accuracy evaluated on PACS; $d=2$.}
\label{tab:results_pacs_d2}
\vskip 0.15in
\begin{center}
\begin{small}
\begin{sc}
\resizebox{0.95\textwidth}{!}{
\begin{threeparttable}
\begin{tabular}{lllllllllllll} 
\toprule
 &  & \multicolumn{5}{c}{$\textit{Acc}_{\text{\textit{id}}}$} &  & \multicolumn{5}{c}{$\textit{Acc}_{\text{\textit{ood}}}$} \\ 
\cmidrule{3-7}\cmidrule{9-13}
 &  & \multicolumn{1}{c}{A} & \multicolumn{1}{c}{C} & \multicolumn{1}{c}{P} & \multicolumn{1}{c}{S} & \multicolumn{1}{c}{$\mu$} &  & \multicolumn{1}{c}{A} & \multicolumn{1}{c}{C} & \multicolumn{1}{c}{P} & \multicolumn{1}{c}{S} & \multicolumn{1}{c}{$\mu$} \\ 
\cmidrule{1-1}\cmidrule{3-7}\cmidrule{9-13}
Central &  & - & - & - & - & \textbf{-} &  & - & - & - & - & \textbf{-} \\
Local &  & 78.1 $\pm$ 3.3 & 69.8 $\pm$ 7.7 & 72.5 $\pm$ 6.0 & 65.5 $\pm$ 7.5 & \textbf{71.5} &  & 34.0 $\pm$ 2.1 & 42.4 $\pm$ 3.8 & 52.7 $\pm$ 7.5 & 44.2 $\pm$ 3.8 & \textbf{43.3} \\ 
\cmidrule{1-1}\cmidrule{3-7}\cmidrule{9-13}
FedAvg &  & 80.9 $\pm$ 1.1 & 74.0 $\pm$ 8.4 & 75.5 $\pm$ 6.7 & 71.7 $\pm$ 5.8 & \textbf{75.5} &  & 42.3 $\pm$ 1.0 & 51.5 $\pm$ 0.3 & 62.8 $\pm$ 1.2 & 56.0 $\pm$ 1.4 & \textbf{53.2} \\
FedProx &  & 80.6 $\pm$ 3.7 & 73.7 $\pm$ 7.8 & 75.7 $\pm$ 6.3 & 70.3 $\pm$ 5.3 & \textbf{75.1} &  & 41.4 $\pm$ 0.8 & 52.8 $\pm$ 1.2 & 62.9 $\pm$ 1.0 & 54.9 $\pm$ 0.3 & \textbf{53.0} \\
pFedHN &  & 77.7 $\pm$ 4.8 & 71.6 $\pm$ 5.8 & 73.9 $\pm$ 5.0 & 65.4 $\pm$ 7.0 & \textbf{72.2} &  & 40.6 $\pm$ 0.6 & 50.8 $\pm$ 1.5 & 58.3 $\pm$ 2.4 & 52.5 $\pm$ 5.1 & \textbf{50.6} \\ 
\cmidrule{1-1}\cmidrule{3-7}\cmidrule{9-13}
FedSR &  & 79.2 $\pm$ 2.7 & 72.4 $\pm$ 8.2 & 74.3 $\pm$ 8.1 & 69.4 $\pm$ 4.1 & \textbf{73.8} &  & 40.9 $\pm$ 0.5 & 50.2 $\pm$ 2.7 & 61.9 $\pm$ 1.2 & 53.5 $\pm$ 2.7 & \textbf{51.6} \\
FedGMA &  & 74.6 $\pm$ 4.8 & 69.7 $\pm$ 7.4 & 71.1 $\pm$ 7.3 & 67.1 $\pm$ 6.4 & \textbf{70.6} &  & 38.6 $\pm$ 1.4 & 48.6 $\pm$ 0.7 & 60.1 $\pm$ 2.1 & 52.5 $\pm$ 3.1 & \textbf{49.9} \\ 
\cmidrule{1-1}\cmidrule{3-7}\cmidrule{9-13}
hFedF &  & 78.2 $\pm$ 2.5 & 74.9 $\pm$ 5.8 & 76.3 $\pm$ 5.2 & 73.9 $\pm$ 4.5 & \textbf{75.8} &  & 44.1 $\pm$ 0.7 & 54.4 $\pm$ 0.8 & 62.3 $\pm$ 1.5 & 58.7 $\pm$ 0.4 & \textbf{54.9} \\
\bottomrule
\end{tabular}
\begin{tablenotes}
\footnotesize
\item[] Abbreviations: $\mu$ - average; A - art painting; C - cartoon; P - photo; S - sketch.
\end{tablenotes}
\end{threeparttable}
}
\end{sc}
\end{small}
\end{center}
\vskip -0.1in
\end{table*}

\begin{table*}[htb]
\caption{Accuracy evaluated on PACS; $d=3$.}
\label{tab:results_pacs_d3}
\vskip 0.15in
\begin{center}
\begin{small}
\begin{sc}
\resizebox{0.95\textwidth}{!}{
\begin{threeparttable}
\begin{tabular}{lllllllllllll} 
\toprule
 &  & \multicolumn{5}{c}{$\textit{Acc}_{\text{\textit{id}}}$} &  & \multicolumn{5}{c}{$\textit{Acc}_{\text{\textit{ood}}}$} \\ 
\cmidrule{3-7}\cmidrule{9-13}
 &  & \multicolumn{1}{c}{A} & \multicolumn{1}{c}{C} & \multicolumn{1}{c}{P} & \multicolumn{1}{c}{S} & \multicolumn{1}{c}{$\mu$} &  & \multicolumn{1}{c}{A} & \multicolumn{1}{c}{C} & \multicolumn{1}{c}{P} & \multicolumn{1}{c}{S} & \multicolumn{1}{c}{$\mu$} \\ 
\cmidrule{1-1}\cmidrule{3-7}\cmidrule{9-13}
Central &  & 80.4 $\pm$ 1.5 & 75.5 $\pm$ 0.9 & 74.9 $\pm$ 1.5 & 70.7 $\pm$ 1.1 & \textbf{75.4} &  & 40.8 $\pm$ 1.2 & 53.1 $\pm$ 1.1 & 65.0 $\pm$ 0.7 & 55.8 $\pm$ 0.5 & \textbf{53.7} \\
Local &  & 75.1 $\pm$ 2.6 & 71.0 $\pm$ 2.8 & 69.3 $\pm$ 1.7 & 62.9 $\pm$ 4.8 & \textbf{69.6} &  & 35.2 $\pm$ 1.9 & 46.6 $\pm$ 3.1 & 55.2 $\pm$ 4.4 & 46.7 $\pm$ 4.5 & \textbf{45.9} \\ 
\cmidrule{1-1}\cmidrule{3-7}\cmidrule{9-13}
FedAvg &  & 81.8 $\pm$ 2.2 & 76.2 $\pm$ 1.9 & 78.1 $\pm$ 2.1 & 71.6 $\pm$ 2.7 & \textbf{76.9} &  & 40.7 $\pm$ 0.5 & 54.1 $\pm$ 1.2 & 63.5 $\pm$ 0.5 & 55.6 $\pm$ 1.9 & \textbf{53.5} \\
FedProx &  & 81.0 $\pm$ 3.3 & 78.2 $\pm$ 1.0 & 76.9 $\pm$ 2.3 & 73.1 $\pm$ 3.3 & \textbf{77.3} &  & 41.1 $\pm$ 0.7 & 54.6 $\pm$ 1.2 & 63.0 $\pm$ 0.7 & 56.3 $\pm$ 0.9 & \textbf{53.8} \\
pFedHN &  & 78.4 $\pm$ 2.2 & 76.0 $\pm$ 2.8 & 72.8 $\pm$ 1.3 & 70.1 $\pm$ 3.4 & \textbf{74.3} &  & 38.4 $\pm$ 0.7 & 55.0 $\pm$ 0.5 & 60.8 $\pm$ 1.7 & 50.0 $\pm$ 2.7 & \textbf{51.1} \\ 
\cmidrule{1-1}\cmidrule{3-7}\cmidrule{9-13}
FedSR &  & 80.4 $\pm$ 1.9 & 75.2 $\pm$ 3.0 & 75.8 $\pm$ 2.1 & 71.8 $\pm$ 4.5 & \textbf{75.8} &  & 38.9 $\pm$ 1.0 & 52.4 $\pm$ 0.8 & 64.4 $\pm$ 0.9 & 53.2 $\pm$ 0.3 & \textbf{52.2} \\
FedGMA &  & 78.2 $\pm$ 3.0 & 74.2 $\pm$ 2.5 & 72.0 $\pm$ 4.5 & 66.6 $\pm$ 3.6 & \textbf{72.7} &  & 38.5 $\pm$ 0.8 & 50.8 $\pm$ 1.6 & 60.7 $\pm$ 1.9 & 49.6 $\pm$ 3.6 & \textbf{49.9} \\ 
\cmidrule{1-1}\cmidrule{3-7}\cmidrule{9-13}
hFedF &  & 80.8 $\pm$ 3.8 & 75.9 $\pm$ 2.7 & 75.2 $\pm$ 2.1 & 72.3 $\pm$ 3.5 & \textbf{76.1} &  & 42.8 $\pm$ 0.5 & 55.6 $\pm$ 0.8 & 64.5 $\pm$ 0.9 & 57.9 $\pm$ 1.1 & \textbf{55.2} \\
\bottomrule
\end{tabular}
\begin{tablenotes}
\footnotesize
\item[] Abbreviations: $\mu$ - average; A - art painting; C - cartoon; P - photo; S - sketch.
\end{tablenotes}
\end{threeparttable}
}
\end{sc}
\end{small}
\end{center}
\vskip -0.1in
\end{table*}

\clearpage
\vfill
\subsubsection{Performance on Office-Home dataset}

Our comprehensive results on Office-Home dataset detailed in \Autoref{tab:results_officehome_d1}, \Autoref{tab:results_officehome_d2}, and \Autoref{tab:results_officehome_d3} demonstrate {hFedF}'s superior domain generalization performance, consistently outperforming benchmarks across all domain complexities.

\begin{table*}[htb]
\caption{Accuracy evaluated on Office-Home; $d=1$.}
\label{tab:results_officehome_d1}
\vskip 0.15in
\begin{center}
\begin{small}
\begin{sc}
\resizebox{0.95\textwidth}{!}{
\begin{threeparttable}
\begin{tabular}{lllllllllllll} 
\toprule
 &  & \multicolumn{5}{c}{$\textit{Acc}_{\text{\textit{id}}}$} &  & \multicolumn{5}{c}{$\textit{Acc}_{\text{\textit{ood}}}$} \\ 
\cmidrule{3-7}\cmidrule{9-13}
 &  & \multicolumn{1}{c}{A} & \multicolumn{1}{c}{C} & \multicolumn{1}{c}{P} & \multicolumn{1}{c}{R} & \multicolumn{1}{c}{$\mu$} &  & \multicolumn{1}{c}{A} & \multicolumn{1}{c}{C} & \multicolumn{1}{c}{P} & \multicolumn{1}{c}{R} & \multicolumn{1}{c}{$\mu$} \\ 
\cmidrule{1-1}\cmidrule{3-7}\cmidrule{9-13}
Central &  & - & - & - & - & - &  & - & - & - & - & - \\
Local &  & 55.1 $\pm$ 10.8 & 44.8 $\pm$ 15.1 & 40.4 $\pm$ 15.1 & 50.5 $\pm$ 17.7 & \textbf{47.7} &  & 11.8 $\pm$ 4.0 & 16.5 $\pm$ 4.5 & 20.8 $\pm$ 8.8 & 19.6 $\pm$ 2.3 & \textbf{17.2} \\ 
\cmidrule{1-1}\cmidrule{3-7}\cmidrule{9-13}
FedAvg &  & 54.3 $\pm$ 7.4 & 46.1 $\pm$ 15.2 & 43.2 $\pm$ 12.2 & 47.6 $\pm$ 19.3 & \textbf{47.8} &  & 19.3 $\pm$ 0.5 & 26.5 $\pm$ 1.1 & 35.9 $\pm$ 0.5 & 34.5 $\pm$ 0.5 & \textbf{29.1} \\
FedProx &  & 54.7 $\pm$ 8.0 & 46.6 $\pm$ 15.2 & 44.8 $\pm$ 10.8 & 48.2 $\pm$ 18.4 & \textbf{48.6} &  & 19.3 $\pm$ 0.5 & 26.4 $\pm$ 0.6 & 36.6 $\pm$ 0.4 & 34.1 $\pm$ 0.8 & \textbf{29.1} \\
pFedHN &  & 57.1 $\pm$ 11.0 & 48.8 $\pm$ 16.0 & 43.9 $\pm$ 13.3 & 49.7 $\pm$ 16.6 & \textbf{49.9} &  & 13.9 $\pm$ 4.2 & 20.3 $\pm$ 4.3 & 26.2 $\pm$ 8.4 & 24.1 $\pm$ 3.5 & \textbf{21.1} \\ 
\cmidrule{1-1}\cmidrule{3-7}\cmidrule{9-13}
FedSR &  & 54.5 $\pm$ 7.2 & 46.5 $\pm$ 14.9 & 44.3 $\pm$ 12.5 & 46.8 $\pm$ 18.0 & \textbf{48.1} &  & 19.3 $\pm$ 0.8 & 27.6 $\pm$ 0.9 & 37.4 $\pm$ 0.9 & 34.3 $\pm$ 0.7 & \textbf{29.6} \\
FedGMA &  & 54.4 $\pm$ 7.4 & 45.7 $\pm$ 14.4 & 43.1 $\pm$ 11.7 & 47.7 $\pm$ 18.8 & \textbf{47.7} &  & 19.5 $\pm$ 0.5 & 26.4 $\pm$ 1.0 & 37.3 $\pm$ 0.4 & 34.6 $\pm$ 0.5 & \textbf{29.5} \\ 
\cmidrule{1-1}\cmidrule{3-7}\cmidrule{9-13}
hFedF &  & 56.8 $\pm$ 8.2 & 47.1 $\pm$ 12.6 & 44.2 $\pm$ 10.8 & 49.6 $\pm$ 16.6 & \textbf{49.4} &  & 20.0 $\pm$ 0.8 & 28.4 $\pm$ 0.6 & 37.4 $\pm$ 1.3 & 33.4 $\pm$ 3.2 & \textbf{29.8} \\
\bottomrule
\end{tabular}
\begin{tablenotes}
\footnotesize
\item[] Abbreviations: $\mu$ - average; A - art; C - clipart; P - product; R - real world.
\end{tablenotes}
\end{threeparttable}
}
\end{sc}
\end{small}
\end{center}
\vskip -0.1in
\end{table*}

\begin{table*}[htb]
\caption{Accuracy evaluated on Office-Home; $d=2$.}
\label{tab:results_officehome_d2}
\vskip 0.15in
\begin{center}
\begin{small}
\begin{sc}
\resizebox{0.95\textwidth}{!}{
\begin{threeparttable}
\begin{tabular}{lllllllllllll} 
\toprule
 &  & \multicolumn{5}{c}{$\textit{Acc}_{\text{\textit{id}}}$} &  & \multicolumn{5}{c}{$\textit{Acc}_{\text{\textit{ood}}}$} \\ 
\cmidrule{3-7}\cmidrule{9-13}
 &  & \multicolumn{1}{c}{A} & \multicolumn{1}{c}{C} & \multicolumn{1}{c}{P} & \multicolumn{1}{c}{R} & \multicolumn{1}{c}{$\mu$} &  & \multicolumn{1}{c}{A} & \multicolumn{1}{c}{C} & \multicolumn{1}{c}{P} & \multicolumn{1}{c}{R} & \multicolumn{1}{c}{$\mu$} \\ 
\cmidrule{1-1}\cmidrule{3-7}\cmidrule{9-13}
Central &  & - & - & - & - & - &  & - & - & - & - & - \\
Local &  & 45.6 $\pm$ 4.7 & 38.4 $\pm$ 8.2 & 34.7 $\pm$ 5.5 & 43.4 $\pm$ 5.1 & \textbf{40.5} &  & 13.7 $\pm$ 2.7 & 18.3 $\pm$ 1.6 & 23.7 $\pm$ 4.9 & 23.2 $\pm$ 2.3 & \textbf{19.7} \\ 
\cmidrule{1-1}\cmidrule{3-7}\cmidrule{9-13}
FedAvg &  & 57.2 $\pm$ 4.4 & 50.3 $\pm$ 6.5 & 44.5 $\pm$ 5.7 & 53.3 $\pm$ 4.6 & \textbf{51.3} &  & 19.2 $\pm$ 0.6 & 26.7 $\pm$ 0.5 & 37.0 $\pm$ 1.0 & 34.9 $\pm$ 0.6 & \textbf{29.4} \\
FedProx &  & 56.8 $\pm$ 3.6 & 49.6 $\pm$ 7.7 & 45.2 $\pm$ 5.8 & 54.4 $\pm$ 3.8 & \textbf{51.5} &  & 19.9 $\pm$ 0.7 & 26.8 $\pm$ 0.3 & 37.0 $\pm$ 0.8 & 35.2 $\pm$ 0.5 & \textbf{29.7} \\
pFedHN &  & 50.2 $\pm$ 4.7 & 43.2 $\pm$ 7.9 & 40.1 $\pm$ 5.4 & 46.7 $\pm$ 8.7 & \textbf{45.1} &  & 16.0 $\pm$ 1.7 & 21.7 $\pm$ 2.7 & 28.9 $\pm$ 4.1 & 27.7 $\pm$ 4.4 & \textbf{23.6} \\ 
\cmidrule{1-1}\cmidrule{3-7}\cmidrule{9-13}
FedSR &  & 56.9 $\pm$ 4.0 & 50.0 $\pm$ 7.0 & 46.4 $\pm$ 5.1 & 52.7 $\pm$ 5.1 & \textbf{51.5} &  & 19.8 $\pm$ 0.3 & 26.6 $\pm$ 0.3 & 36.9 $\pm$ 1.1 & 35.7 $\pm$ 0.6 & \textbf{29.8} \\
FedGMA &  & 57.4 $\pm$ 3.3 & 50.1 $\pm$ 6.8 & 44.6 $\pm$ 5.0 & 53.4 $\pm$ 4.3 & \textbf{51.4} &  & 20.1 $\pm$ 0.0 & 27.4 $\pm$ 0.3 & 37.2 $\pm$ 0.6 & 35.4 $\pm$ 0.3 & \textbf{30.0} \\ 
\cmidrule{1-1}\cmidrule{3-7}\cmidrule{9-13}
hFedF &  & 56.4 $\pm$ 4.0 & 52.1 $\pm$ 6.6 & 46.1 $\pm$ 4.3 & 54.0 $\pm$ 5.6 & \textbf{52.1} &  & 20.4 $\pm$ 1.0 & 28.5 $\pm$ 1.1 & 38.8 $\pm$ 0.6 & 37.2 $\pm$ 0.6 & \textbf{31.2} \\
\bottomrule
\end{tabular}
\begin{tablenotes}
\footnotesize
\item[] Abbreviations: $\mu$ - average; A - art; C - clipart; P - product; R - real world.
\end{tablenotes}
\end{threeparttable}
}
\end{sc}
\end{small}
\end{center}
\vskip -0.1in
\end{table*}

\clearpage
\vfill
\subsubsection{Performance on VLCS dataset}

Our comprehensive results on VLCS dataset detailed in \Autoref{tab:results_vlcs_d1}, \Autoref{tab:results_vlcs_d2}, and \Autoref{tab:results_vlcs_d3} demonstrate {hFedF}'s superior domain generalization performance, consistently outperforming benchmarks across all domain complexities.

\begin{table*}[htb]
\caption{Accuracy evaluated on VLCS; $d=1$.}
\label{tab:results_vlcs_d1}
\vskip 0.15in
\begin{center}
\begin{small}
\begin{sc}
\resizebox{0.95\textwidth}{!}{
\begin{threeparttable}
\begin{tabular}{lllllllllllll} 
\toprule
 &  & \multicolumn{5}{c}{$\textit{Acc}_{\text{\textit{id}}}$} &  & \multicolumn{5}{c}{$\textit{Acc}_{\text{\textit{ood}}}$} \\ 
\cmidrule{3-7}\cmidrule{9-13}
 &  & \multicolumn{1}{c}{C} & \multicolumn{1}{c}{L} & \multicolumn{1}{c}{S} & \multicolumn{1}{c}{V} & \multicolumn{1}{c}{$\mu$} &  & \multicolumn{1}{c}{C} & \multicolumn{1}{c}{L} & \multicolumn{1}{c}{S} & \multicolumn{1}{c}{V} & \multicolumn{1}{c}{$\mu$} \\ 
\cmidrule{1-1}\cmidrule{3-7}\cmidrule{9-13}
Central &  & - & - & - & - & \textbf{-} &  & - & - & - & - & \textbf{-} \\
Local &  & 62.4 $\pm$ 3.0 & 71.6 $\pm$ 17.4 & 70.5 $\pm$ 17.3 & 75.1 $\pm$ 15.4 & \textbf{69.9} &  & 60.4 $\pm$ 12.4 & 41.5 $\pm$ 14.5 & 36.5 $\pm$ 12.5 & 36.2 $\pm$ 10.3 & \textbf{43.6} \\ 
\cmidrule{1-1}\cmidrule{3-7}\cmidrule{9-13}
FedAvg &  & 57.8 $\pm$ 4.1 & 67.6 $\pm$ 10.6 & 70.3 $\pm$ 14.8 & 66.7 $\pm$ 10.4 & \textbf{65.6} &  & 73.8 $\pm$ 1.6 & 54.8 $\pm$ 0.5 & 54.5 $\pm$ 1.3 & 48.7 $\pm$ 1.7 & \textbf{58.0} \\
FedProx &  & 56.3 $\pm$ 3.8 & 67.5 $\pm$ 13.4 & 68.6 $\pm$ 15.5 & 69.2 $\pm$ 10.8 & \textbf{65.4} &  & 73.0 $\pm$ 1.9 & 53.3 $\pm$ 0.8 & 51.7 $\pm$ 0.7 & 46.8 $\pm$ 0.8 & \textbf{56.2} \\
pFedHN &  & 58.2 $\pm$ 8.7 & 65.1 $\pm$ 20.2 & 63.3 $\pm$ 23.9 & 67.2 $\pm$ 19.3 & \textbf{63.4} &  & 70.0 $\pm$ 5.4 & 48.0 $\pm$ 8.0 & 41.9 $\pm$ 8.5 & 40.2 $\pm$ 6.9 & \textbf{50.0} \\ 
\cmidrule{1-1}\cmidrule{3-7}\cmidrule{9-13}
FedSR &  & 59.9 $\pm$ 3.8 & 65.9 $\pm$ 12.5 & 69.8 $\pm$ 15.0 & 68.2 $\pm$ 11.4 & \textbf{66.0} &  & 74.4 $\pm$ 0.5 & 54.4 $\pm$ 0.5 & 51.1 $\pm$ 1.1 & 49.2 $\pm$ 0.4 & \textbf{57.2} \\
FedGMA &  & 59.4 $\pm$ 4.0 & 68.0 $\pm$ 13.4 & 69.0 $\pm$ 16.2 & 69.3 $\pm$ 10.1 & \textbf{66.4} &  & 73.6 $\pm$ 1.6 & 55.0 $\pm$ 0.8 & 51.8 $\pm$ 1.6 & 47.5 $\pm$ 0.6 & \textbf{57.0} \\ 
\cmidrule{1-1}\cmidrule{3-7}\cmidrule{9-13}
hFedF &  & 58.4 $\pm$ 3.3 & 68.4 $\pm$ 16.8 & 68.2 $\pm$ 16.6 & 69.4 $\pm$ 13.8 & \textbf{66.1} &  & 72.8 $\pm$ 0.7 & 52.6 $\pm$ 2.4 & 53.9 $\pm$ 0.3 & 48.4 $\pm$ 1.7 & \textbf{56.9} \\
\bottomrule
\end{tabular}
\begin{tablenotes}
\footnotesize
\item[] Abbreviations: $\mu$ - average; C - Caltech101; L - LabelMe; S - SUN09; V - PASCAL VOC.
\end{tablenotes}
\end{threeparttable}
}
\end{sc}
\end{small}
\end{center}
\vskip -0.1in
\end{table*}

\begin{table*}[htb]
\caption{Accuracy evaluated on VLCS; $d=2$.}
\label{tab:results_vlcs_d2}
\vskip 0.15in
\begin{center}
\begin{small}
\begin{sc}
\resizebox{0.95\textwidth}{!}{
\begin{threeparttable}
\begin{tabular}{lllllllllllll} 
\toprule
 &  & \multicolumn{5}{c}{$\textit{Acc}_{\text{\textit{id}}}$} &  & \multicolumn{5}{c}{$\textit{Acc}_{\text{\textit{ood}}}$} \\ 
\cmidrule{3-7}\cmidrule{9-13}
 &  & \multicolumn{1}{c}{C} & \multicolumn{1}{c}{L} & \multicolumn{1}{c}{S} & \multicolumn{1}{c}{V} & \multicolumn{1}{c}{$\mu$} &  & \multicolumn{1}{c}{C} & \multicolumn{1}{c}{L} & \multicolumn{1}{c}{S} & \multicolumn{1}{c}{V} & \multicolumn{1}{c}{$\mu$} \\ 
\cmidrule{1-1}\cmidrule{3-7}\cmidrule{9-13}
Central &  & - & - & - & - & \textbf{-} &  & - & - & - & - & \textbf{-} \\
Local &  & 57.1 $\pm$ 3.4 & 62.4 $\pm$ 5.0 & 64.2 $\pm$ 7.1 & 69.2 $\pm$ 5.3 & \textbf{63.2} &  & 63.4 $\pm$ 7.0 & 50.0 $\pm$ 3.0 & 44.2 $\pm$ 5.7 & 42.2 $\pm$ 4.7 & \textbf{50.0} \\ 
\cmidrule{1-1}\cmidrule{3-7}\cmidrule{9-13}
FedAvg &  & 60.3 $\pm$ 1.8 & 65.4 $\pm$ 4.4 & 66.7 $\pm$ 6.4 & 70.0 $\pm$ 5.2 & \textbf{65.6} &  & 74.1 $\pm$ 1.0 & 56.0 $\pm$ 1.0 & 51.7 $\pm$ 0.4 & 49.1 $\pm$ 0.9 & \textbf{57.7} \\
FedProx &  & 59.3 $\pm$ 2.7 & 65.6 $\pm$ 5.2 & 65.4 $\pm$ 5.4 & 68.2 $\pm$ 4.6 & \textbf{64.6} &  & 73.4 $\pm$ 1.2 & 54.8 $\pm$ 0.5 & 50.7 $\pm$ 1.3 & 48.0 $\pm$ 1.8 & \textbf{56.7} \\
pFedHN &  & 60.0 $\pm$ 3.2 & 63.5 $\pm$ 4.8 & 63.9 $\pm$ 6.8 & 68.5 $\pm$ 6.0 & \textbf{63.9} &  & 72.3 $\pm$ 2.9 & 54.6 $\pm$ 0.5 & 52.7 $\pm$ 2.1 & 48.5 $\pm$ 0.4 & \textbf{57.0} \\ 
\cmidrule{1-1}\cmidrule{3-7}\cmidrule{9-13}
FedSR &  & 60.4 $\pm$ 2.9 & 66.3 $\pm$ 5.4 & 64.7 $\pm$ 6.3 & 69.4 $\pm$ 4.8 & \textbf{65.2} &  & 74.5 $\pm$ 1.0 & 53.2 $\pm$ 0.4 & 52.6 $\pm$ 0.7 & 49.6 $\pm$ 2.0 & \textbf{57.4} \\
FedGMA &  & 61.2 $\pm$ 3.3 & 65.7 $\pm$ 3.5 & 66.8 $\pm$ 7.7 & 68.6 $\pm$ 5.7 & \textbf{65.6} &  & 72.1 $\pm$ 1.4 & 55.6 $\pm$ 0.6 & 51.4 $\pm$ 0.8 & 47.5 $\pm$ 0.6 & \textbf{56.7} \\ 
\cmidrule{1-1}\cmidrule{3-7}\cmidrule{9-13}
hFedF &  & 61.2 $\pm$ 2.1 & 68.2 $\pm$ 4.6 & 66.9 $\pm$ 8.1 & 71.1 $\pm$ 5.3 & \textbf{66.8} &  & 74.8 $\pm$ 1.2 & 55.0 $\pm$ 1.5 & 53.1 $\pm$ 0.5 & 49.6 $\pm$ 1.0 & \textbf{58.1} \\
\bottomrule
\end{tabular}
\begin{tablenotes}
\footnotesize
\item[] Abbreviations: $\mu$ - average; C - Caltech101; L - LabelMe; S - SUN09; V - PASCAL VOC.
\end{tablenotes}
\end{threeparttable}
}
\end{sc}
\end{small}
\end{center}
\vskip -0.1in
\end{table*}

\end{document}